%% file: main.tex
\documentclass[10pt,twocolumn,letterpaper]{article}
\usepackage{cvpr}

\usepackage{times}
\usepackage{epsfig}
\usepackage{graphicx}
\usepackage{amsmath}
\usepackage{amssymb}

\usepackage[pagebackref=true,breaklinks=true,colorlinks,bookmarks=false]{hyperref}

\usepackage{amssymb} %checkmark

\usepackage{enumitem}
\usepackage{booktabs}

\usepackage{amsmath}
\usepackage{xspace}

\usepackage[dvipsnames,table]{xcolor}
\usepackage[pagebackref=true,breaklinks=true,colorlinks,bookmarks=false]{hyperref}

\usepackage{cleveref}
\definecolor{darkergreen}{RGB}{21, 152, 56}
\definecolor{red2}{RGB}{252, 54, 65}
\newcommand\redp[1]{\textcolor{red2}{(#1)}}
\newcommand\greenp[1]{\textcolor{darkergreen}{(#1)}}

\definecolor{lightblue}{rgb}{0,0,1}

\usepackage{arydshln}

\usepackage{pifont}
\usepackage[table]{xcolor}
\usepackage{algorithm}
\usepackage{algorithmic}

\newcommand*\colorcmark[1]{%
  \expandafter\newcommand\csname #1cmark\endcsname{\textcolor{#1}{\ding{51}}}%
}
\colorcmark{green}

\newcommand*\colorxmark[1]{%
  \expandafter\newcommand\csname #1xmark\endcsname{\textcolor{#1}{\ding{55}}}%
}
\colorxmark{red}

\usepackage{multirow}
\usepackage{enumitem}
\usepackage[table]{xcolor}
\usepackage{amsthm}
\newlength\savewidth

    \newlength\thinwidth
    \definecolor{Gray}{gray}{0.92}
    \newcolumntype{a}{>{\columncolor{Gray}}c}
    \definecolor{LightCyan}{rgb}{0.88,1,1}
    \definecolor{lightblue}{rgb}{0,0,1}
    \definecolor{darkergreen}{RGB}{21, 152, 56}
    \definecolor{highlightRowColor}{gray}{0.92}

\crefname{section}{Sec.}{Secs.}
\Crefname{section}{Section}{Sections}
\Crefname{table}{Table}{Tables}
\crefname{table}{Tab.}{Tabs.}

\usepackage{bm}

\definecolor{LightCyan}{rgb}{0.92,1,1}
\definecolor{darkergreen}{RGB}{21, 152, 56}
\definecolor{red2}{RGB}{252, 54, 65}

\definecolor{pcolor}{HTML}{CA2C92}
\definecolor{ocolor}{HTML}{6A5ACD}
\definecolor{citecolor}{HTML}{0071BC}
\definecolor{linkcolor}{HTML}{ED1C24}

\def \ModelName          {\textit{PTP}\xspace}
\def \ModelNameVILT        {\textit{PTP-}ViLT\xspace}
\def \ModelNameCLIP          {\textit{PTP-}CLIP\xspace}
\def \ModelNameBLIP          {\textit{PTP-}BLIP\xspace}
\definecolor{demphcolor}{RGB}{144,144,144}
\newcommand{\demph}[1]{\textcolor{demphcolor}{#1}}
\def \GlobalTableRescale {1}

\begin{document}

\title{Position-guided Text Prompt for Vision-Language Pre-training}

\author{Alex Jinpeng Wang$^{2}$ \quad Pan Zhou$^{1}\;$ \quad Mike Zheng Shou$^{2}$\quad Shuicheng Yan$^{1}$\\
$^1$Sea AI Lab\quad $^2$Show Lab, National University of Singapore\quad \\}
\maketitle

%%%%%%%%% ABSTRACT
\begin{abstract}

Vision-Language Pre-Training (VLP) has shown promising capabilities to align image and text pairs, facilitating a broad variety of cross-modal learning tasks. However, we observe that VLP models often lack the visual grounding/localization capability which is critical for many downstream tasks such as visual reasoning. In this work, we propose a novel Position-guided Text Prompt (\ModelName)  paradigm to enhance the visual grounding ability of cross-modal models trained with VLP. Specifically, in the VLP phase,  \ModelName divides the image into $N\times N$ blocks, and identifies the objects in each block through the widely used object detector in VLP.  It then reformulates the visual grounding task into a fill-in-the-blank problem given a  \ModelName  by encouraging the model to predict the objects in the given blocks or regress the blocks of a given object, e.g. filling ``[\textcolor{pcolor}{P}]"  or ``[\textcolor{ocolor}{O}]" in a \ModelName
``The block [\textcolor{pcolor}{P}] has a [\textcolor{ocolor}{O}]".
This mechanism improves the visual grounding capability of VLP models and thus helps them better handle various downstream tasks.
By introducing \ModelName into several state-of-the-art VLP frameworks, we observe consistently significant improvements across representative cross-modal learning model architectures and several benchmarks,  e.g. zero-shot Flickr30K Retrieval (+4.8 in average recall@1) for ViLT \cite{vilt} baseline, and COCO Captioning (+5.3 in CIDEr) for SOTA BLIP \cite{blip} baseline.
Moreover, \ModelName  achieves comparable results with object-detector based methods \cite{uniter,oscar,vinvl}, and much faster inference speed since \ModelName discards  its object detector for   inference while the later cannot.
Our code and pre-trained weight will be released at \url{https://github.com/sail-sg/ptp}.
   
\end{abstract}

%%%%%%%%% BODY TEXT

%%%%%%%%% BODY TEXT
\vspace{-0.2em}
\section{Introduction}
\label{sec:intro}
\vspace{-0.2em}
\input{Figures/1_Introduction}

% para1: the history of pre-training, first pretraining and then fine-tuning
The vision-and-language pre-training (VLP) models like CLIP \cite{CLIP}, ALIGN \cite{align} and CoCa \cite{coca} have greatly advanced the state-of-the-art performance of many cross-modal learning tasks, e.g.,  visual question answering \cite{vqa}, reasoning \cite{nlvr2}, and image captioning \cite{cococaptioning,nocaps}.
Typically, a generic cross-modal model is first pre-trained on large-scale image-caption data in a self-supervised fashion to see sufficient data for better generalization ability, and then fine-tuned on downstream tasks for  adaptation. 
With remarkable effectiveness, 
this pre-training-then-fine-tuning paradigm of VLP models has dominated the multi-modality field.
%

% para2: the importance of position

In VLP, visual grounding is critical for many tasks as observed in previous research \cite{butd,filip}.
%quanhong: not any explanation of Fig content in main text?
To model the position information, traditional VLP models \cite{butd,oscar,vinvl} (the top of Fig.~\ref{fig:1_motivation} (a)) employ a faster-rcnn \cite{fasterrcnn} pre-trained on the 1600 classes Visual Genome \cite{VisualGenome} to extract salient region features and bounding boxes. 
Then these models use both the bounding box and object feature as input. 
In this way, these models not only learn what objects are contained in the salient region and where are these objects.
%However, the limitations are still evident in downstream tasks undergo a slow extraction process.
However, when using region features as input, the model pays attention to the items inside the bounding boxes and ignores the contextual data outside of them \cite{soho}.
More seriously,  on downstream task,
these methods still need to use detectors to extract objects, giving very slow inference speed.
%xtracting the salient region features and bounding boxes are very slow, which hinders these models' application in various downstream tasks.
%However, these method for various downstream task, since extracting the salient region features and bounding boxes are very slow.
%However, when fine-tuning on downstream tasks, these methods often have to disregard  the heavy visual embedder \pz{you do not mention visual embedder? what is it} which indeed impair the performance~\pz{give citations}, since extracting the salient region features and bounding boxes are very slow.   
% In addition, the transfer ability of these models  is limited by the predefined visual vocabulary \cite{vilt} \pz{your method has the same issue? If yes, why do you mention it?}.

To get rid of region feature for higher efficiency,  recent works \cite{vilt,soho} 
%try to get rid of region feature and train VLP models end-to-end.
%ViLT \cite{vilt}, made the first attempt to 
(the middle of Fig.~\ref{fig:1_motivation} (a)) adopt raw-pixel image as input instead of region features, and  train the model with Image Text Matching~\cite{uniter} and Masked Language Modeling \cite{bert} loss end-to-end.
%without  fine-grained alignment (i.e.~one-to-one alignment among objects and text phrases).  
%Despite the running time is ten of times faster than region featured based VLP models, we find that the position information is missed in these models. 
Despite their faster speed, these models  cannot well  learn the object positions and also their relations.  
As shown in Fig. \ref{fig:1_motivation} (b), we observe that a well-trained ViLT model  \cite{vilt} well know what objects are in an image.
But this model does not learn the object positions  accurately. 
For example, it wrongly predicts   ``\textit{the dog is on the right of this image}".  
However, during fine-tuning, downstream tasks actually require the object position information to comprehensively understand the image.
Such a gap largely impairs the performance  on downstream tasks.

In this work, we aims to ease the position missing problem for these end-to-end models, and keep fast inference time for downstream tasks at the same time.
Inspired by the recently prompt learning methods \cite{raffel2020exploring,liu2021pre,colorprompt,maple}, we propose a novel and effective  \textbf{Position-guided Text Prompt (\ModelName)} paradigm (the bottom of Fig.~\ref{fig:1_motivation} (a)) for cross-modality model pre-training.  
The key insight is that by adding position-based co-referential markers in both image and text, visual grounding can be reformulated into a fill-in-the-blank problem, maximally simplify the learning of object information. 
To ground natural language expressions in image data, \ModelName contains two components: (1) block tag generation to divide image into $N \times N$ blocks and to identify object in each block, and (2) text prompt generation that puts the query text into a position-based text  query template. 
%\pz{it is not clear what problem you want to solve ? object and text alignment? If Yes, the region based methods have done. What are the differences? What issues do the region based methods have? how do you solve speed and location together?}

By bringing the position information into pre-training, our \ModelName enables strong visual grounding capabilities of VLP models. 
At the same time, as we do not used object detector for downstream tasks, we keep fast inference time.
Experimental results show that our method outperforms their counterparts by a large margin especially for zero-shot setting. 
For example, our \ModelName-BLIP achieves 3.4\% absolute accuracy gain over CoCa \cite{coca} in zero-shot retrieval Recall@1 on coco dataset with much less training data (4M vs. 3B) and a much smaller model (220M vs. 2.1B).
%achieves 4.2\% absolute accuracy improvement on average with visual reasoning evaluation over BLIP \cite{blip} baseline. 
In addition to the zero-shot task, we show that \ModelName  can achieve strong performance for object position guided visual reasoning and the other common VLP tasks such as visual question answering, and image captioning.

\vspace{-0.2em}
\section{Related Work}
\vspace{-0.2em}
\subsection{Vision-language Pre-training Models}
\vspace{-0.2em}
Existing VLP models can be roughly grouped into three categories according to their architectures:  one-stream models, dual-stream models and dual-stream + fusion encoder model.
All three architectures are introduced below:

\textit{1) One-stream Model}
(e.g., UNITER \cite{uniter}, ViLT \cite{vilt})  in Fig.~\ref{fig:3_2_framework} (a) operates on a concatenation of image and text inputs.
\textit{2) Dual-stream Model} (e.g., CLIP \cite{CLIP}) in  Fig.~\ref{fig:3_2_framework} (b)  uses separate but equally expensive transformer encoders for each modality. 
The two modalities are not concatenated at the input level and interaction between the pooled image vector and text vector at shallow layer. 
\textit{3) Dual-stream with Fusion Model} (e.g., BLIP \cite{blip}) Fig. \ref{fig:3_2_framework} (c) is a combination of one-stream and dual-stream model.

In this work, without loss of generality, we focus on prompting all these three kinds of VLP models due to their prevalence and adaptability to different downstream tasks.
%while applying \ModelName to other VLP models is also applicable.

\input{Figures/3_2_Framework_Variations}

\vspace{-0.1em}
\subsection{Prompt Learning for Computer Vision}
\vspace{-0.1em}
Prompt learning is originally designed for probing knowledge in pre-trained language models to specific downstream tasks~\cite{raffel2020exploring,liu2021pre}. 
Recent years have seen a rise in the study of prompt tuning on vision taks, e.g. multi-modal learning and image understanding. 
The pioneer Color Prompt \cite{colorprompt} adds color prompt on image and text color description for visual grounding. 
Most related to our work is Multi-modality Prompt \cite{maple} which presents multi-modality prompt tuning for VLPT models, achieving promising results on some vision-language tasks. 

However, these efforts, like earlier NLP research, concentrate on prompt engineering in fine-tuning while leaving the pre-training phase unaffected.
The goal of using the prompt design in this work, in contrast, is to provide the model the ability to understand semantic concepts at a finer level while it is still in the pre-training stage.

\vspace{-0.1em}
\subsection{Learn Position Information in VLP}
\vspace{-0.1em}
The grounding ability  has shown to be essential for multiple cross-modality tasks \cite{glip,loctex}.
To introduce this ability into VLP models, bottom-up and top-down \cite{butd} and its follow-up works \cite{oscar,uniter} concatenate  region feature and bounding box vector together.
%Oscar \cite{oscar} introduce object tags and learn to align in object-level.
But object extraction is time-consuming in inference for downstream task.
Recently, some works \cite{glip,loctex,xvlm} propose train the VLP models with additional object localization loss or word patch alignment loss which, however, are hard to extend because they are specifically designed for particular frameworks. In contrast, we aim to propose a general framework for learning position information.
To this end, we propose a simple text prompt that can be plug into existing frameworks easily.

\section{Position-guided Text Prompt}
\vspace{-0.3em}
In this section, we first elaborate on our proposed  Position-guided Text Prompt paradigm (\ModelName for short). Then we  introduce  how to incorporate it with current vision-language pre-training (VLP) frameworks for boosting their visual grounding capabilities by taking the classical and popular VILT~\cite{vilt}, CLIP~\cite{CLIP}  and BLIP~\cite{blip} as examples.

\vspace{-0.3em}
\subsection{\textbf{\ModelName} Paradigm}
\vspace{-0.3em}
%In this work, t
To enhance the visual grounding ability of cross-modal models trained by VLP,  we propose a novel and effective Position-guided Text Prompt (\ModelName) that helps a cross-modal model perceive objects,  and also align these objects with pertinent text.  \ModelName differs from the conventional vision language alignment methods, e.g. \cite{butd,uniter,oscar,vinvl}, that concatenate object feature and bounding box together as input to learn the alignment between objects and pertinent text, and thus paves an alternative way which indeed enjoys some advantages as shown and discussed in Sec.~\ref{sec:3.2}.  As illustrated  in Fig. \ref{fig:3_ppl}, \ModelName has two steps: 1) block tag generation which divides  an input image into several blocks and  also identifies the objects in each block;  and 2) text prompt generation that  reformulates the visual grounding task into a fill-in-the-blank problem according to the object position information in step 1).  Based on these  steps, one can easily plug PTP into a VLP model by solving fill-in-the-blank problem in \ModelName.  We will introduce these two steps below.

\vspace{-0.4em}
\subsubsection{Block Tag Generation} \label{taggeneration}
\vspace{-0.4em}
As shown  in Fig. \ref{fig:3_ppl},  for each image-text pair in the training phase, we evenly divide the input image into $N \times N$ blocks.  
Then we identify  the object  in each block by one of the following two ways:  

\textbf{(1) Object Detector.} 
We first adopt a strong Faster-rcnn \cite{fasterrcnn} used in VinVL \cite{vinvl} to extract all objects for each image.
This  Faster-rcnn version is based on ResNeXt152 and is trained on 1600-classes Visual Genome \cite{VisualGenome}. 
Then we select top-$K$ objects denoted by $\mathcal{O}=\{o_i\}_{i=1}^{K}$ with highest prediction confidence, where 
$o_i=(z_i, q_i)$ denotes an  object with 4-dimensional region position vector  $z$  and object category $q$. 
For each block, we select the objects whose region center are in that block.
At last, the final block tag for this block is $q$ of these selected objects.
In this work, we generate object tag with object detector as default.
% For each object $v_i$, we then compute the center of it according to $z$ and position label $P \in [1,N^2]$ is assigned based on which block the center falls under.
%The center of the chosen object is then determined based on position $z$, and position label $P \in [1,9]$ is assigned based on which block the center falls under. \pz{not clear how to determine $q$}

% Notice that the object detector is only used in the pre-training stage.
% For the downstream evaluation, there is no additional supervision or computation cost.

\input{Figures/3_Main_PPL}

\textbf{(2) CLIP Model.}
Instead of heavy object detector, some recent works \cite{denseclip,regionclip} also try to generate region supervision based on CLIP \cite{CLIP} because of its efficiency and effectiveness. 
Inspired by these works, \ModelName can also generate block-wise object supervision via CLIP (ViT-B) model \footnote{https://huggingface.co/openai/clip-vit-base-patch16}.  
First, we extract $M$ (3000 in default) key words/phrases that are most frequent on the whole text corpus \footnote{Extract key word/phrase with NLTK (/https://github.com/nltk/nltk)}.
These key words/phrases are regarded as our vocabulary $V$.
Then we extract the text feature $e_i, i\in[1,\dots,M]$ of all these $M$ key words/phrases embedding via CLIP text encoder.

Additionally, we take the image embedding $h$ from each block and compute the similarity across every text feature. 
The keyword/phrase with the highest similarity score is selected as the final object tag for this particular block.
Formally, the index of object tag per block is computed as
\begin{equation}
	I = \mbox{argmax}_{y\in[1,\dots,M]} \ \left(\frac{\exp(h^T e_y)}{\sum_{w\in V}^{}\exp(h^T e_w)}\right),
\end{equation}
where $h$ is the visual feature embedding of selected block. Comparing with object detector, the CLIP model have two advantages.  Firstly, as opposed to pre-defined object categories, more diverse object tags are produced. 
%\pz{do you have some experiments to support}  
Secondly, the generation of block tag is much faster than object detector, e.g. 40$\times$ faster than Faster-RCNN (ResNeXt152) model. Please refer to Sec. \ref{sec:ablation} for comparison.

%Compared to object detectors, .
%By choosing the most often occurring objects from the entire corpus, as opposed to object detectors that employ preset, pre-defined object categories, more diverse object tags are produced, supporting open-vocabulary.

\vspace{-0.5em}
\subsubsection{Text Prompt Generation}
\vspace{-0.5em}
For the input image of each training pair,  Sec.~\ref{taggeneration} already generate the object tags and positions which allows us to design  a simple text prompt as follows: 
\begin{equation*}
``\textit{\textit{The block [\textcolor{pcolor}{P}] has a [\textcolor{ocolor}{O}]}.}"
\end{equation*}
where \textcolor{pcolor}{$P$} $\in \{1, \cdots, N^2\}$ denotes the index of selected block and is used to denote the object position; \textcolor{ocolor}{$O$} denotes the object tag generated for the block \textcolor{pcolor}{$P$}. Note, we explore more prompt design choices in Section \ref{sec:ablation}.  
For a certain \textcolor{pcolor}{$P$}, we may have various options for \textcolor{ocolor}{$O$} because the block may contain multiple objects.
%\pz{please introduce how do you solve this issue}. 
For such situation, we select one \textcolor{ocolor}{$O$} at random for each time.
In this way, each sentence in our \ModelName  incorporates fine-grained object position and language into a model, and thus provides a new way to align the objects and pertinent text. 

\subsection{Pre-training with \ModelName}

\label{sec:3.2}
In this work, we integrate our \ModelName into mainstream VLP frameworks, leading to \ModelNameVILT~\cite{vilt}, \ModelNameCLIP~\cite{CLIP}  and \ModelNameBLIP~\cite{blip}.
Following receipt of the \ModelName, we have two options for training these models:

\textbf{Integrate into existing tasks.}
The simplest method for using text prompt is to change the text input.
As shown in Fig. \ref{fig:3_ppl}, the prompted text and original caption were simply padded together.
Formally, the input caption $x$ of our method is represented as:
\begin{equation}
x = [w, q],
\end{equation}
where $w$ is text and $q$ is our generated text prompt.
Then we train the VLP models end-to-end with conventional objectives.
Following   \cite{blip,vilt,CLIP}, we employ Language Modeling (LM) loss, Image-text Matching (ITM), and Image-text Contrastive (ITC) loss for our \ModelName-BLIP;
we use ITM and Masked Language Modeling (MLM) loss to train our \ModelNameVILT; we only use ITC loss to train our \ModelNameCLIP.
We use this method as default for all experiments because of its good performance.

\textbf{As a new pretext task.}
Alternatively, we explore the position prediction as an additional language modeling task. 
Formally, if $D$ is the pretraining data and $y_1, \dots, y_T$ is a training token sequence of our generated text prompt $q$, then at the timestep $t$, we devise our model to predict a probability distribution $p(t)=p(*|y_{1},\dots,y_{t-1})$.
Then we regressively try to maximize the probability of being the correct token. 
The object prediction loss is computed as follow:

\begin{equation}
\mathcal{L}_{\mathrm{PTP}}(\theta)=-\mathbb{E}_{\mathbf{y} \sim D}\left[\sum\nolimits_{t=1}^T \log P_\theta\left(\mathbf{y}_t \mid \mathbf{y}_{<t}\right)\right],
\end{equation}
where $\theta$ is the trainable parameters of the model.
In this way, the model is asked to predict \textit{which block \textcolor{pcolor}{$P$} has objects} and \textit{what object \textcolor{ocolor}{$O$} is in this block}.

\textbf{Discussion.}
Notably, our method does not need to modify the base network and can be applied to any VLP models without bells and whistles.
The model is designed to learn position information from raw-pixel image.
Note that only during the pre-training stage, we would require the object's position information; yet on downstream tasks, we evaluate model in normal end-to-end ways without object information to get rid of the heavy object feature extraction.

\section{Experiments}
In this section, we empirically evaluate \ModelName on multiple downstream tasks and present a comprehensive study.

\input{Tables/tab1_zero_shot_retrieval}
\input{Tables/tab2_retrieval.tex}

\subsection{Experimental Settings}
We first describe the pre-training experimental conditions, including the datasets, training configurations, evaluation procedures, and baseline models used in our studies.

\textbf{Datasets.}
As in earlier studies \cite{oscar,vinvl}, we begin by using a 4M setup made up of four popular pre-training datasets (COCO \cite{coco}, VG \cite{VisualGenome}, SBU \cite{sbu} and CC3M \cite{cc3m}).
Following recent work \cite{blip}, we also explore 14M setting, which includes additional CC12M \cite{cc12m} (actually only 10M image urls available) dataset besides 4M datasets.
We refer readers to supplementary material for more dataset details.

\textbf{Training Settings}.
Our models are implemented in PyTorch \cite{pytorch} and pre-trained on 8 NVIDIA A100 GPUs.
For the optimizer and training hyperparameter, we follow the original implementation in baseline works for fair comparison.
For image augmentation, we explore RandAugment \cite{randaugment} and use all the original policies except for color inversion since color information is important.
We augment the bounding box in same way as image for affine transformation like rotation.
We take random image crops of resolution $224 \times 224$ during pre-training, and increase the image resolution to $384 \times 384$ for finetuning.

\textbf{Baselines.} 
We evaluate three variants of pre-training frameworks, including one-stream ViLT \cite{vilt}, dual-encoder CLIP \cite{CLIP}, and fusion-encoder BLIP \cite{blip}, for their superior performance.
For fair comparisons, we adopt the ViT-B/16~\cite{vit} as base vision encoder and use same dataset.

\subsection{Main Results}
In this section, we integrated our \ModelName into existing networks and compare to existing VLP methods on a wide range of vision-language downstream tasks.
Then we introduce each task and finetuning strategy.
More details can be found in the supplementary material.

\input{Tables/tab3_captioning}

\subsubsection{Image-Text Retrieval}
We evaluate \ModelName for both image-to-text retrieval (TR) and text-to-image retrieval (IR) on COCO and Flickr30K benchmarks. 
For \ModelName-BLIP, following original implementation, we adopt an additional re-ranking strategy. 

We first report zero-shot retrieval result on both image-to-text and text-to-image setting in Tab. \ref{tab:zero-shot-retrieval-table}. 
We find \ModelName significantly improves baselines on all metrics.
For example, for ViLT \cite{vilt} baseline, \ModelName leads to 13.8 \% absolute improvement (from 41.3 \% to 55.1 \%) over Recall@1 of image to text retrieval on MSCOCO.
In addition, based on strong BLIP \cite{blip}, our \ModelName-BLIP even outperforms CoCa \cite{coca} on most recalls of MSCOCO with much less data.

A summary comparison about fine-tuned setting between different models appears in Tab. \ref{tbl:results:finetuned_retrieval}, from which we observe that: 
(1) \ModelName outperforms the BLIP and ViLT baselines by a large margin in both datasets.
For example, \ModelName-ViLT achieves an impressive 5.3\%  improvement on R@1 of TR in MSCOCO. 
(2) 
With strong BLIP as baseline,  \ModelName-BLIP leads to state-of-the-art performance at same scale.
Notice that the training cost remains the same BLIP baseline, because we train \ModelName with the same settings as the baseline and do not increase the maximum input text token.
We can even reduce the gap between 4M setting and ALBEF \cite{albef} (14M data), with similar framework.

From all these results above, we point out UNITER \cite{uniter}, OSCAR~\cite{oscar}, VinVL \cite{vinvl}, ImageBERT \cite{imagebert} all use faster-rcnn as we used.
However, our \ModelName leads to much better results than these related works.
Besides, we only use object detector in pre-training stage.
This indicates \textit{object detector is not the secret for success and how to leverage the position information is essential important for VLP models}.

\subsubsection{Image Captioning}
This task asks the model to describe the input image.
We consider two datasets for image captioning: No-Caps \cite{nocaps} and COCO \cite{coco}, both evaluated using the model finetuned on COCO with the LM loss.
Similar to BLIP, we start each caption with the phrase ``a picture of," which yields marginally better results.
We do not pre-train using the COCO dataset to avoid information leakage.
For No-Caps dataset, following BLIP, we adopts a zero-shot setting (evaluate directly with the captioning model trained on CoCo dataset).

As shown in Tab.~\ref{tab:caption}, related works utilizing a comparable quantity of pre-training data perform significantly worse than \ModelName-BLIP.
The results of our method are closed to the VinVL~\cite{vinvl} with fewer training samples and smaller image.
Finally, with 14M setting, our method leads to close result with LEMON, which trained on billions data and requires two times higher resolution image.

\input{Tables/tab4_vqa_nlvr2}

\subsubsection{Visual Question Answering}
VQA \cite{vqa} requires the model to predict an answer given an image and a question. 
For \ModelName-ViLT, we formulating VQA as a multi-answer classification task. 
For \ModelName-BLIP, we follow \cite{albef,blip} and consider it as an answer generation task that allows open-vocabulary VQA for better result.
%This is shown to improve VQA accuracy.

The results are reported in Tab. \ref{tab:vqa_nlvr}. 
Compared to ViLT baseline, \ModelName brings 1.8\% gains on both dev split.
With 14M setting, \ModelName-BLIP achieves better performance than SimVLM \cite{simvlm}, which uses 1.8B training samples and a ViT-Large based vision backbone.

\subsubsection{Visual Reasoning}

Natural Language Visual Reasoning (NLVR$^2$) \cite{nlvr2} task is a binary classification task given triplets of two images and a question in natural language.
This task relies on position information heavily.
As shown in Tab. \ref{tab:vqa_nlvr}, SimVLM \cite{simvlm} is outperformed by \ModelNameBLIP, which has a reasonable model size and was pretrained on fewer instances.
Meanwhile, our method is also closed to VinVL$_{large}$ model that adopt larger model and use object feature from strong object detector instead of raw-pixel image as input.

\input{Tables/tab5_msrvtt_retrieval}

\subsubsection{Video-Language Tasks}
We analyze the generalization ability of our method to video-language tasks in this experiment. 
Specifically, we perform zero-shot transfer to text-to-video retrieval in Tab. \ref{tab:zsl_video_retrieval}, where we directly evaluate the models trained on COCO-retrieval. 
We just uniformly sample 8 frames each video in order to process video input, then concatenate the frame features into a single sequence.
Our method leads to better result than  OA-Trans \cite{oatrans} that focus on retrieval task, which showcase the generality capability of \ModelName.
%Also, note that this simple approach ignores all temporal information.

\subsection{Ablation \& Design Choices}
\label{sec:ablation}

In this section, we first evaluate our method on retrieval task over three well-known baselines under 4M setting for comparison.
Then we train a BLIP model on CC3M as baseline and perform various ablations.

\subsubsection{The Variations of Architecture.}
We experiment with three distinct kind baselines: ViLT, CLIP, and BLIP in order to explore the impact of \ModelName.
Tab. \ref{tbl:abl_architectures} reports the performance on the COCO 5K test set.
Comparing the outcomes of these baseline experiments, we find that \ModelName greatly improves the i2t and t2i performance.
This suggests that \ModelName has good generality.

In addition, we also compare the running time.
Since we do not use object detector or prompt in downstream task, the computation cost keep consistent with baseline models but 20 times faster than object feature based VinVL \cite{vinvl}.

\input{Tables/tab10_abl_architevtures}

\subsubsection{Text Prompt vs. Additional Pretext Task}
We examine the effects of regarding \ModelName as a new pretext task.
In this way, the pretext task does not influence the other pre-training objectives, such as ITM and ITC, but it does add to the cost of computation. 
Contrarily, the prompt design simply modifies the text input, therefore it will have an impact on all pre-training objectives.

\input{Tables/tab9_pretext_vs_prompt}

We report the result in Tab. \ref{tab:pretext_vs_prompt}.
We observe both Pretext and Prompt design improved the baseline over all four tasks.
However, prompting is far preferable to pretext, particularly for COCO captioning CIDER (127.2 vs 123.5). 
In this work, we use prompt as default due to its efficiency.

\subsubsection{Other Types of Text Prompt}

In this experiment, we explore six different kind of prompts:
\emph{i}. The [\textcolor{ocolor}{O}] is in block [\textcolor{pcolor}{P}].
\emph{ii}. The block [\textcolor{pcolor}{P}] looks like  [\textcolor{ocolor}{O}].
\emph{iii}. The [\textcolor{ocolor}{O}] is in which block? In [\textcolor{pcolor}{P}].
\emph{iv}. The [\textcolor{ocolor}{O}] is located in block [\textcolor{pcolor}{P}]. 
\emph{v}. ($X_1$, $Y_1$, $W$, $H$) has a [\textcolor{ocolor}{O}]. $(X_1,Y_1)$ is the top left point and $W,H$ are the width and height for bounding box.
\emph{vi}. The block [\textcolor{pcolor}{P}] has a [\textcolor{ocolor}{O}].
\emph{vii}. The block [NP] has a [\textcolor{ocolor}{O}]. NP means we use nouns to represent the block position. e.g, from upper left to bottom right.
More variations can be found in the supplementary.

We report the result in Tab. \ref{tab:abl_prompt_design} and observe precise position does not produce superior results to block, the reason maybe precise position is hard to learn.
In addition, we find use block ID (like 0) or nouns (like upper left) remain similar results.
In the end, we discover that the hybrid version does not produce the best outcomes.

\input{Tables/tab6_prompt_abl}

\input{Tables/tab7_object_pred_var}

\subsubsection{The Importance of Position in Text Prompt}
In this experiment, we examine the efficacy of prompting our \ModelName for information at various granularities, such as without Positional.
We simply use \textit{[\textcolor{pcolor}{P}] has [\textcolor{ocolor}{O}]} when remove prompt.
We list the results in Tab. \ref{tab:abl_object_pred}. 
We observe:
\emph{i}. It's interesting to see that each component is crucial. 
Without any one component, the downstream performance to get progressively poorer.
\emph{ii}. Although OSCAR \cite{oscar} discovered that using object tags as a supplementary input improved results when area features were used as input, we have shown that object tags are ineffective when raw pixel images are used. 
This serves as an illustration of the need to create a workable prompt for understanding the alignment between object tags and image region.

\subsubsection{Number of Blocks}
We explore if more fine-grained position information helps in our \ModelName.
In Fig. \ref{fig:ablation_patch_num}, we varying the number of blocks from $1 \times 1$ (remove position information in \ModelName) to $4 \times 4$ and report the relative performance based on both BLIP and ViLT models.
As can be seen, the results for both backbones are improved when the number of blocks is more than 1. 
However, once there are 16 blocks, all downstream activities experience a relative drop in performance.
The reason may be that the predicted bounding box deviates from the localization of the real object, resulting in a mesh that is too small and may not contain the selected object.
We hence recommend using $3\times3$ blocks, as it enjoys accurateness.

\input{Tables/tab8_remove_detector}

\subsubsection{Is Object Detector Necessary? }
In this work, a part of predicted bounding box information is coming from Faster-rcnn \cite{fasterrcnn}.
In order to verify the expressive power of object, we also consider two variations:
\emph{i}. Pure clip similarity. This design choice is adapted mainly for efficiency reasons, where utilizing object detector is time consuming and not easy to access sometimes.
\emph{ii}. In addition to the powerful ResNext152-based object detector, we also use a smaller Faster-rcnn network that utilizes ResNet101 as backbone.

\input{Figures/4_Abl_Patch_Num}

The results are reported in Tab. \ref{tab:remove_detector}.
We also report the overall feature extracting time on 8 NVIDIA V100 GPUs.
As can be seen from the table, we found that using stronger detector leads to better result, but bring huge computation cost at the same time.
Moreover, we observe the result of CLIP embedding is very closed to Faster-rcnn (ResNeXt152).
In addition, it takes only around 2.3\% time of Faster-rcnn (ResNeXt152) version to extract pseudo label for each grid.
We came to the conclusion that a clip model is a good alternative of object detector in \ModelName.

\subsection{Visualization}
\label{sec:4_visulaize}

To explore whether model training with the \ModelName framework does indeed learn position information, we design a fill-in-the-blank evaluation experiment in this section.
Follow ViLT \cite{vilt}, we masked some key words and asked the model to predict the masked words and show its corresponding heatmap.
We design two text prompts, given the noun to predict the localization and given the localization to predict the missing noun.
We show top-3 predictions and more visualization results can be found in supplementary.

The results are shown in Fig. \ref{fig:block_pred}.
On the one hand, we find that the \ModelName-ViLT can make correct object prediction based on the block position information and its visual concepts.
On the other hand, when only masked the position information, we witness a high predicted probability value for corrected block.
For example, in the bottom of Fig. \ref{fig:block_pred}, our model find all patches looks like \textit{``man''} correctly.
Based on these experiments and Fig. \ref{fig:1_motivation}, we conclude that the \ModelName can help the base VLP model learn position information very well based on our simple text prompt.

\input{Figures/4_Block_Pred_Visualize}

\input{Figures/4-Token_Visualize}
Furthermore, we cluster the token-level features with K-Means algorithm for ViLT and \ModelName-ViLT.
Intuitively, the token with similar semantic should be clustered together.
We show the visualization result in Fig. \ref{fig:token_visualize}.
Comparing with ViLT baseline, we observe that our method can cluster similar patches more accurate.
This illustrate our \ModelName have fairly accurate learns semantic information.

% \textcolor{blue}{Efficiency Analysis}

% more experiments about efficiency of PTP

\section{Limitations and Conclusion}
We first try to leverage the position information from existing object detector/trained model to VLP models with simple prompt.
We provide a success practice cross-modal prompt settings to aid prompt engineering.
Through rigorous experiments, we showed that \ModelName could serve as a general-purpose pipeline and improve the learning of position information without much extra computation cost.
However, at this time, \ModelName does not take into account how to deal with the wrong object tag.
Additionally, this work does not adequately explore more complicated prompts.
Future research will also examine how well \ModelName performs on additional vision-language tasks.

{\small
\bibliographystyle{ieee_fullname}
\bibliography{main}
}

%\clearpage

\appendix
\section*{Appendix}

\section{Pre-training and Fine-tuning Details}
\subsection{Statistics of the Pre-training Datasets}
\input{Supplementary/tab1_dataset_statistic}

In this work, we explore both 4M and 14M setting.
The 14M setting is a combination of 4M setting and CC-12M.
We report the data statistics in Tab. \ref{tab:data}.
Since these URLs are from Interent and a part of them already invalid, we only download 2.8M data of CC3M and 10.2M data of CC12M dataset, correspondingly.
Notice that BLIP baseline use 3M data for CC3M, which is slightly more than our version.
The amount of images that containing bounding box for CC3M  is 2.69M, and for CC12M is 7M.
These bounding boxs are used in our \ModelName.
For quick evaluation, we pre-train the BLIP model for 50K steps rather than the 200K steps used in earlier works~\cite{vilt,xvlm}.

\input{Supplementary/tab3_ft_details}

\input{Supplementary/tab2_more_prompt}

\subsection{Hyper-parameters for Downstream Tasks}

We first report the hyper-parameters of BLIP baseline in Tab. \ref{tabs:detailed_vl}.
The final decoder outputs from the encoder-decoder model BLIP can be used for multimodal understanding and generation.
Thus, we evaluate on popular vision-language benchmarks. 
We mainly follow the same setup introduced in BLIP~\cite{blip}. 
The optimizer for all task is AdamW~\cite{adamw}.
We only train the retrieval task for 6 epochs in order to increase efficiency, and we think that more epochs will produce better results.

For ViLT~\cite{vilt} baseline, we evaluate mainly on three tasks: vision-question answering, image-text retrieval and natural language visual reasoning.
The hyper-parameters for ViLT on downstream tasks are reported in Tab. \ref{tabs:detailed_vl}.
For the CLIP baseline, we use the same hyper-parameters setting as BLIP baseline.

\input{Supplementary/figure1_block_mask}

\section{More Ablation Study}
\subsection{More prompt design}
We also explore multiple other prompt design choices in this section.
The model is trained on CC3M and we evaluate on three downstream tasks.
Specifically, we exploring the following ways:
\emph{i}. \textbf{Multiple Tags.} We observe that a block may contain many objects for many cases.
We try to refine the text prompt as \textit{The block [P] has objects [O$_1$], [O$_2$] and [O$_3$]}.
Keep in mind that each block has a different object number.
\emph{ii}. \textbf{Multiple Position.} 
We created a multiple position setup taking into account that one object could appear in numerous blocks.
In practical, we refine the prompt as a question-answer pairs.
\emph{iii}. \textbf{Synonymous Substitution} We replace ``block'' with ``region'' and ``is'' with ``looks like''.

The results is reported in Tab. \ref{tab:more_prompt}.
We observe that the multiple objects or multiple position not helps the model's performance on downstream tasks too much, we also observe that the language modeling loss is higher than baseline.
This shows that the assignment is too challenging for the model to learn.
We also see that the outcome of simple synonymous replacement maintains consistency with the outcome of the original text prompt. 
We find that modeling location information only requires a straightforward prompt.

\input{Supplementary/figure2_vary_objects}
\subsection{How many Objects do we need?}
To generate object tag, we use Faster-RCNN as default and detect at least 10 objects from each image.
In this experiment, we varied the number of items from 5 to 30, exploring the effects of various object counts.

The result is shown in Fig. \ref{fig:vary_object}.
We observe there exist a slightly rising trend at the beginning of BLIP baseline.
This demonstrates how crucial data diversity is for activities that come afterwards. 
The findings, however, are poorer when there are more objects. 
The cause is because a large number of objects with low confidence simultaneously produce false predictions.
In this work, we set the object number as 10 as default.

\input{Supplementary/tab4_part_object}

\subsection{Part Bounding Box Annotation}
As some urls for CC dataset already invalid and some images have wrong fromat, we extract objects from  2.7M data of CC3M and 7M data of CC12M.
In this way, only 10M of pre-training sample have objects available.
We also report the result with 14M setting.
Specifically, we only use original text without text prompt if we do not have object available.

The result is shown in Tab. \ref{tab:part_object}.
We observe 68.6\% result in 134.6 CiDER value on COCO Captioning and 82.9 on NLVR accuracy.
This illustrates more annotated samples leads to better result.
This also encourage \textit{PTP} is suitable for large-scale pre-training.

% \subsection{Video Question Answering}
% In this section, we transfer our trained model to zero-shot Video Question Answering.

\section{More Visualization}

% \subsection{Visualization about left/right..}
% We show the attention weight and the corresponding predicted words.
% We observe. 
% Replace left/right .

\subsection{Bounding Box Visualization}
In this section, we show object detection result with our generated text prompt.
Specifically, we random select one object from $V$ and then we visualize the original image and its bounding box's mask.
Notice we augment these bounding box as the same as original image for affine transformation.

We random select some samples from the overall dataset and the result is reported in Fig. \ref{fig:block_mask}.
We also observe the bounding box maybe very large and cross multiple blocks in some examples (e.g. the first case in the third row).
Since we use RandAugment~\cite{randaugment} in this work, some object may be outside of the broder of input image.
For such situation, we just replace the specific position with [X], the final \ModelName is \textit{The block [X] has a [O]}.
We also find that some masks may be no square.
For example, the last example in the third row.

\input{Supplementary/figure3_downstream_visualization}

\subsection{Case Analysis}
In this experiment, we show some cases about position information in Fig. \ref{fig:samples_visualization}.
The position information is important for various downstream tasks on the top.

Then, since a large of samples in VQA tasks include position information usually. 
We ask our model to do the vqa tasks and select some representative samples.
Specifically, we show the prediction probability and predicted nouns in the bottom of this figure.
We observe that the \ModelName give accurate prediction on most cases, which illustrates our \ModelName learns position information better.

\end{document}

%% file: Figures/1_Introduction.tex
\begin{figure}[t]
  \centering
    \includegraphics[width=\linewidth]{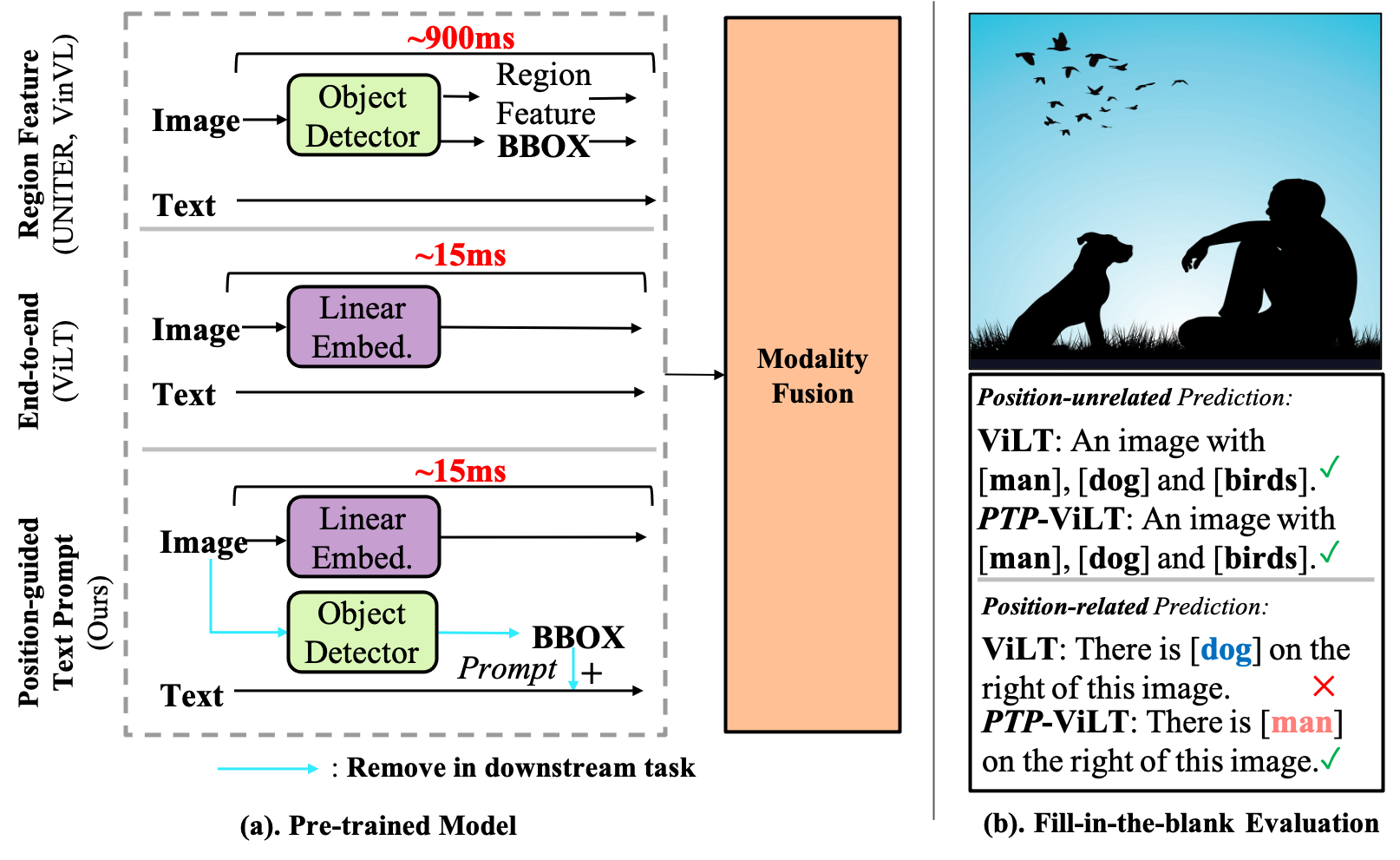}
       \vspace{-1.4em}
   \caption{
   {Comparison of three VLP learning frameworks  and their performance}. (a) compares  region feature based VLP (RF-VLP), end-to-end VLP (E2E-VLP), and our position-guided text prompt based VLP (PTP-VLP). Our PTP-VLP only needs about 15ms for inference which is the same as E2E-VLP but is much faster than RF-VLP. (b) On position-aware questions widely occurred in many downstream tasks, with  masked text and image input, RF-VLP and PTP-VLP can well predict objects, while E2E-VLP cannot pinpoint the position information of the object  in the image.  
%   Position information is essential for many downstream tasks (a) but missed in a well-trained ViLT \cite{vilt} model (b).}
%     Given input masked text and images, we ask the model to predict masked objects.
%     The ViLT model accurately predicts objects, but it is unable to pinpoint where each object is in the image.
   }
   \vspace{-1.2em}
   \label{fig:1_motivation}
\end{figure}

%% file: Figures/3_2_Framework_Variations.tex
\begin{figure}[t]
  \centering
    \includegraphics[width=.9\linewidth]{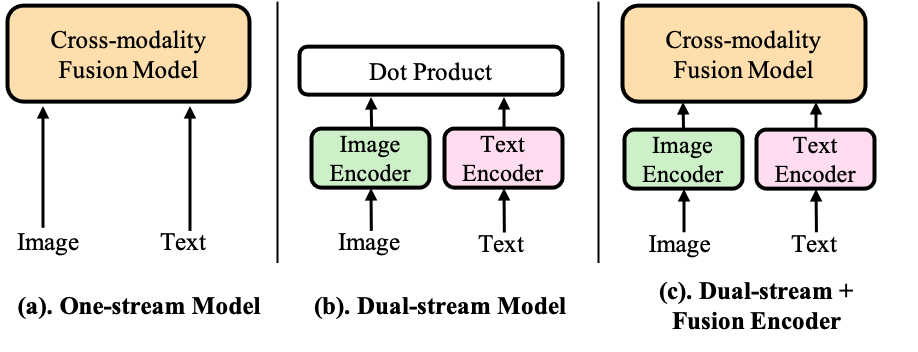}
\vspace{-1em}
  \caption{
  \textbf{Three widely-used categories of vision-and-language models.}
  The main difference is where to perform cross-modality information fusion.
  One-stream fuse at early stage and dual-stream fuse at late stage, while the last type fuse at middle stage.
  }
  \vspace{-1.3em}
  \label{fig:3_2_framework}
\end{figure}

%% file: Figures/3_Main_PPL.tex
\begin{figure}[t]
  \centering
    \includegraphics[width=.8\linewidth]{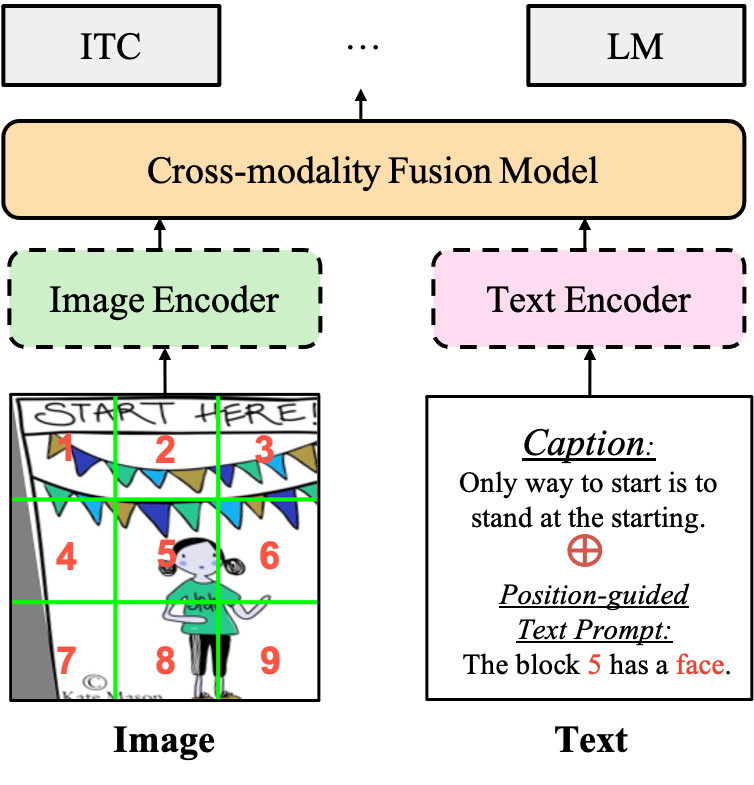}
    \vspace{-1em}
  \caption{
  \textbf{
 Overall framework. 
  }
    Any pre-training framework (one-stream, dual-stream, dual-stream+fusion encoder in Fig.~\ref{fig:3_2_framework}) and most objectives  can be integrated with our \ModelName. 
    Dashed line indicates that the model may not exist. 
    %for different architectures.
    We remove the text prompt for the downstream task and evaluate the model as usual. 
  }
  \vspace{-1em}
  \label{fig:3_ppl}
\end{figure}

%% file: Tables/tab1_zero_shot_retrieval.tex
\begin{table*}
%\vspace{-3mm}
\center
\small
\caption{
\textbf{Results of zero-shot image-text retrieval on Flickr30K and MSCOCO datasets.}
We gray out the methods that train on much larger corpus or use much larger models.
$\dag$ means the model implemented by ourself and trained on same dataset since the original datasets is not accessible or not trained on these splits.
The Avg is the mean of all image-to-text recalls and text-to-image recalls.
}
\resizebox{\linewidth}{!}{
\begin{tabular}{lll|ccccccl|ccccccl}
\toprule
 \bf Method & \bf \#Images & \bf Parameters & \multicolumn{7}{c}{\bf MSCOCO (5K test set)} & \multicolumn{7}{c}{\bf Flickr30K (1K test set)} \\
&&& \multicolumn{3}{c}{Image $\rightarrow$ Text} & \multicolumn{3}{c}{Text $\rightarrow$ Image} && \multicolumn{3}{c}{Image $\rightarrow$ Text} & \multicolumn{3}{c}{Text $\rightarrow$ Image} \\
\cmidrule(lr){4-6} \cmidrule(lr){7-9} \cmidrule(lr){11-13} \cmidrule(lr){14-16}
 &&  & R@1 & R@5 & R@10 & R@1 & R@5 & R@10 & Avg & R@1 & R@5  & R@10  & R@1 & R@5 & R@10 & Avg \\
\midrule
Unicoder-VL~\cite{unicoder}&4M &170M &$-$   & $-$  & $-$& $-$  &$-$  & $-$ && 64.3  & 85.8& 92.3 &48.4&76.0&85.2 & 75.3 \\
ImageBERT \cite{imagebert} &4M & 170M & 44.0 & 71.2          & 80.4   & 32.3 & 59.0 & 70.2 & 59.5& 70.7          & 90.2          & 94.0  & 54.3 & 79.6 & 87.5 & 79.4       \\
ViLT \cite{vilt} &4M& 87M & 41.3 & 79.9 & 87.9 & 37.3 & 67.4 & 79.0& 65.5 &69.7 & 91.0 & 96.0 &53.4&80.7&88.8 & 79.9 \\
\bf \ModelName-ViLT (ours) & 4M  & 87M& 55.1 & 82.3 & 89.1 & 43.5 & 70.2 & 81.2 & 70.2$_{+4.7}$&74.5 & 93.7 & 96.5 &60.3&85.5&90.4 & 83.5$_{+3.6}$ \\
BLIP \dag \cite{blip} &4M &220M & 57.4 & 81.1 & 88.7 & 41.4 & 66.0 & 75.3 &68.3& 76.0 & 92.8 &  96.1 &58.4&80.0&86.7&81.7\\
\bf \ModelName-BLIP (ours) &4M&220M& \bf 69.7 & \bf 90.0 & \bf 95.7 & \bf 49.5 & \bf 75.9 & \bf 84.2 & \bf 77.3$_{+9.0}$& \bf 86.4 & \bf 97.6 & \bf 98.9 &\bf 67.0&\bf 87.6&\bf 92.6 & \bf 88.4$_{+6.7}$\\
\midrule
\bf \ModelName-BLIP (ours) &14M&220M& 71.4 &  91.3 & 95.5 & 51.2  & 77.4& 87.1  &78.6&  87.1 & 98.4 &  99.3 & 73.1& 91.0& 94.8 & 90.3\\
\midrule
\rowcolor{LightCyan}
\demph{CLIP} \cite{CLIP} &\demph{300M}&\demph{173M}&\demph{58.4}&\demph{81.5}&\demph{88.1}& \demph{37.8} & \demph{62.4} & \demph{72.2} &\demph{66.7}& \demph{88.0} & \demph{98.7} & \demph{99.4} &\demph{68.7}&\demph{90.6}&\demph{95.2} & \demph{90.1} \\
\rowcolor{LightCyan}
\demph{ALIGN} \cite{align} &\demph{1.8B}& \demph{820M}& \demph{58.6}& \demph{83.0} & \demph{89.7} & \demph{45.6} & \demph{69.8} & \demph{78.6} &\demph{70.9}& \demph{88.6} & \demph{98.7}& \demph{99.7} &\demph{75.7}&\demph{93.8}&\demph{96.8} & \demph{92.2} \\
\rowcolor{LightCyan}
\demph{FILIP} \cite{filip} &\demph{340M}&\demph{787M}& \demph{61.3} & \demph{84.3} & \demph{90.4} & \demph{45.9} & \demph{70.6} & \demph{79.3} &\demph{72.0}& \demph{89.8} & \demph{99.2} & \demph{99.8} &\demph{75.0}&\demph{93.4}&\demph{96.3} & \demph{92.3}\\
\rowcolor{LightCyan} 
\demph{Flamingo} \cite{flamingo} & \demph{2.1B} &\demph{80B}& \demph{65.9} & \demph{87.3} & \demph{92.9} & 
\demph{48.0} & \demph{73.3} & \demph{82.1} &\demph{74.9}& \demph{89.3} & \demph{98.8} & \demph{99.7} & \demph{79.5} & \demph{95.3} & \demph{97.9} & \demph{93.4} \\
\rowcolor{LightCyan} 
\demph{CoCa} \cite{coco} & \demph{3B} &\demph{2.1B}& \demph{66.3} & \demph{86.2} & \demph{91.8} & 
\demph{51.2} & \demph{74.2} & \demph{82.0} & \demph{75.3}& \demph{92.5} & \demph{99.5} & \demph{99.9} & \demph{80.4} & \demph{95.7} & \demph{97.7} & \demph{94.3} \\
%\rowcolor{LightCyan}
%&\demph{BEIT-3} \cite{beit3} &\demph{36M+Corpus}& \demph{94.9} &\demph{99.9}&\demph{100.0}&- & - &- \\
\bottomrule       
\end{tabular}
}
\label{tab:zero-shot-retrieval-table}
%\vspace{-3mm}
\end{table*}

%% file: Tables/tab2_retrieval.tex
\begin{table*}[t]
\centering
\small
\caption{
\textbf{Finetuning results of image-to-text retrieval and text-to-image retrieval on COCO and Flickr30K.}
Notice that UNITER \cite{uniter}, OSCAR~\cite{oscar} and VinVL \cite{vinvl} all use bounding box and object feature.
BeIT-3~\cite{beit3} uses additional 160GB text corpus.
%We gray out the method that utilize very large pre-training data.
}
\resizebox{\textwidth}{!}{%
\begin{tabular}{lll|ccccccl|ccccccl}
\toprule
\multicolumn{1}{l}{\bf Method} &
\bf \#Images & \multicolumn{1}{l}{\bf Parameters} 
&\multicolumn{7}{c}{\bf MSCOCO (5K test set)} & \multicolumn{7}{c}{\bf Flickr30K (1K test set)} \\
 &&& \multicolumn{3}{c}{Image $\rightarrow$ Text} & \multicolumn{3}{c}{Text $\rightarrow$ Image} && \multicolumn{3}{c}{Image $\rightarrow$ Text} & \multicolumn{3}{c}{Text $\rightarrow$ Image} \\
 \cmidrule(lr){4-6} \cmidrule(lr){7-9} \cmidrule(lr){11-13} \cmidrule(lr){14-16}&
 && R@1 & R@5 & R@10 & R@1 & R@5 & R@10 & Avg& R@1 & R@5 & R@10 & R@1 & R@5 & R@10 & Avg \\
\midrule
UNITER \cite{uniter} & 4M &155M& 65.7 & 88.6 & 93.8 & 52.9 & 79.9 & 88.0 &78.2& 87.3 & 98.0 & 99.2 & 75.6 & 94.1 & 96.8 &91.8 \\
OSCAR \cite{oscar} & 4M &155M& 70.0 & 91.1 & 95.5 & 54.0 & 80.8 & 88.5 & $-$ & $-$ & $-$ & $-$ & $-$ & $-$& $-$  \\
%VILLA \cite{villa} & 4M && $-$ & $-$ & $-$ & $-$ & $-$ & $-$ && 87.9 & 97.5 & 98.8 & 76.3 & 94.2 & 96.8 \\
VinVL \cite{vinvl} & 4M&157M& 74.6 & 92.6 & 96.3 & 58.1 & 83.2 & 90.1 &82.5& $-$ & $-$ & $-$ & $-$ & $-$ & $-$& $-$ \\
ViLT \cite{vilt} & 4M &87M& 61.8 & 86.2 & 92.6 & 41.3 & 72.0 & 82.5 &72.7 & 81.4 & 95.6 & 97.6 & 61.9 & 86.8 & 92.8 & 86.0 \\
\bf  \ModelName-ViLT (ours) & 4M&87M& 67.1 & 90.5 & 94.3 & 45.3 & 79.1 & 88.4 &77.5$_{+4.8}$& 85.2 & 96.9 & 98.5 & 68.8 & 91.4 & 95.3 & 89.4$_{+3.4}$ \\
BLIP $\dag$ \cite{blip} & 4M  &220M& 75.2 & 93.3 &  96.3  &  57.4 &  82.1 & 89.5 &82.3 &94.0 & 99.1 & 99.7 & 82.5 & 96.4 & 98.2 & 95.0 \\
 \bf \ModelName-BLIP (ours) & 4M &220M& \bf 77.6 & \bf 94.2 &  \bf 97.0  & \bf 59.4 & \bf 83.4 & \bf 90.4 &\bf 83.7$_{+1.4}$& \bf 96.1 & \bf 99.8 & \bf 100.0 & \bf 84.2 & \bf 96.6 & \bf 98.6 & \bf 95.9$_{+0.9}$ \\
\midrule
ALBEF \cite{albef} & 14M &210M & 77.6 & 94.3 & 97.2 & 60.7 & 84.3 & 90.5 &84.1 & 95.9 & 99.8 & 100.0 & 85.6 & 97.5 & 98.9 & 96.3 \\
BLIP \cite{blip} & 14M &220M & 80.6 & 95.2 & 97.6 & 63.1 & 85.3 & 91.1 &85.5 &96.6 & 99.8 & 100.0 & 87.2 & 97.5 & 98.8 & 96.7 \\
 \bf \ModelName-BLIP (ours) & 14M &220M &  81.5   &  95.9  & 97.9 & 64.9 & 87.4  & 92.2 & 86.6$_{+1.1}$ & \bf 97.0 & \bf  99.9 &  \bf 100.0 &  \bf 87.7 & \bf  98.2 & \bf 99.3 & \bf 97.0$_{+0.3}$\\
\midrule
\rowcolor{LightCyan}
\demph{ALIGN} \cite{align} & \demph{1.8B} &\demph{820M}& \demph{77.0} & \demph{93.5} & \demph{96.9} & \demph{59.9} & \demph{83.3} & \demph{89.8} &\demph{83.4}& \demph{95.3} & \demph{99.8} & \demph{100.0} & \demph{84.9} & \demph{97.4} & \demph{98.6} & \demph{96.0} \\
\rowcolor{LightCyan}
\demph{FILIP} \cite{filip} & \demph{340M} &\demph{787M}& \demph{78.9} & \demph{94.4} & \demph{97.4} & \demph{61.2} & \demph{84.3} & \demph{90.6} &\demph{84.5}& \demph{96.6} & \demph{100.0} & \demph{100.0} & \demph{87.1} & \demph{97.7} & \demph{99.1} & \demph{96.8} \\
\rowcolor{LightCyan}
\demph{Florence} \cite{florence} & \demph{900M}&\demph{893M}& \demph{81.8} & \demph{95.2} & \demph{$-$} & \demph{63.2} & \demph{85.7} & \demph{$-$} &\demph{$-$}& \demph{97.2} & \demph{99.9} & \demph{$-$} & \demph{87.9} & \demph{98.1} & \demph{$-$} & \demph{$-$} \\
\rowcolor{LightCyan}
\demph{Beit-3}~\cite{beit3} & \demph{35M+} & \demph{1.9B} & \demph{84.8} & \demph{96.5}& \demph{98.3}& \demph{67.2}& \demph{87.7} & \demph{92.8} & \demph{87.8}& \demph{98.0}&\demph{100.0}&\demph{100.0}&\demph{90.3}&\demph{98.7}&\demph{99.5} & \demph{97.7}\\
\bottomrule
\end{tabular}
}
\vspace{-3mm}
\label{tbl:results:finetuned_retrieval}
\end{table*}

%% file: Tables/tab3_captioning.tex
\begin{table*}[h]
	\centering	
		\caption
	{
	%\small	
	\textbf{Comparison with state-of-the-art image captioning methods on NoCaps and COCO Caption.}
	%All methods optimize the cross-entropy loss during finetuning. 
	C: CIDEr, S: SPICE, B@4: BLEU@4.	
	Notice that VinVL\ddag~and LEMON\ddag~require high resolution (800$\times$1333) input images. 
	%SimVLM$_\mathrm{huge}$ uses 13$\times$ more training data and a larger vision backbone than ViT-L.
	}
	\resizebox{\textwidth}{!}{%
	\begin{tabular}	{l l  l |  l l l l l l l l | l l l l }
		\toprule	 	
	  \bf \multirow[t]{3}{*}{Method}& \bf \multirow[t]{3}{*}{{\bf \#Images}} & \bf Parameters &  \multicolumn{8}{c|}{\bf NoCaps validation} & 
	 \multicolumn{4}{c}{\bf COCO Caption}\\
	 && & \multicolumn{2}{c}{in-domain} & \multicolumn{2}{c}{near-domain} & \multicolumn{2}{c}{out-domain} & \multicolumn{2}{c|}{Overall} & \multicolumn{4}{c}{Karpathy test}\\
	 && & CIDEr & SPICE & CIDEr & SPICE & CIDEr & SPICE & CIDEr & SPICE & B@4 &METEOR & SPICE & CIDEr\\
	  \midrule
	   %VL-T5/BART & & 35.1 & 116.6 & - & - \\
	   %ViLT~\cite{vilt}  & 70.94 & - & 75.24 & 76.21& - & - \\
	   OSCAR~\cite{oscar} & 4M &155M&79.6 & 12.3 & 66.1 & 11.5 & 45.3 & 9.7 & 80.9 & 11.3 & 37.4 & 30.7 & 23.5 & 127.8 \\
	   VinVL\ddag~\cite{vinvl} & 5.7M &347M& 103.1 & 14.2 & 96.1 & 13.8 & 88.3 & 12.1 & 95.5 & 13.5 &  38.5 & 30.4 & 23.4 & \bf 130.8 \\
  	   BLIP \dag \cite{blip}  &4M &220M& 106.5& 14.4 & 99.3 & 13.6 & 95.6 & 13.0 & 98.8 & 14.2  &37.0 & $-$ &$-$ & 122.6 \\
	   \bf \ModelName-BLIP  (ours) &4M &220M& \bf 108.3 & \bf 14.9&\bf 105.0 &\bf 14.2&\bf 105.6 & \bf 14.2 &\bf 106.0 & \bf 14.7  & \bf 38.6 & 30.3 & 23.3 &  128.9 \\
	   \midrule
       Enc-Dec~\cite{cc12m} & 15M &$-$& 92.6 & 12.5 & 88.3 & 12.1 & 94.5 & 11.9 & 90.2 & 12.1 & $-$ &$-$&$-$&110.9\\
	   BLIP \cite{blip} &14M &220M& 111.3 & 15.1 & 104.5 & 14.4 & 102.4 & 13.7  & 105.1 & 14.4  &38.6 & $-$ & $-$& 129.7 \\
   	   \bf \ModelName-BLIP (ours) &14M &220M &\bf 112.8 & \bf 15.2 & \bf 107.3 & \bf 14.9 & \bf 108.1 & \bf 14.3 & \bf 106.3 & \bf 14.7  & \bf 40.1 & \bf 30.4 & \bf 23.7& \bf 135.0 \\
	   \midrule
     \rowcolor{LightCyan}
   	   \demph{SimVLM$_\mathrm{huge}$} \cite{simvlm} & \demph{1.8B}& \demph{1.2B} & \demph{113.7} & \demph{$-$} & \demph{110.9} & \demph{$-$} & \demph{115.2} & \demph{$-$} & \demph{112.2} & \demph{$-$} & \demph{40.6} & \demph{33.7} & \demph{25.4}& \demph{143.3}\\	
   	    \rowcolor{LightCyan}
    \demph{LEMON$_\mathrm{huge}$}\ddag~ \cite{lemon}& \demph{200M} &\demph{675M}& \demph{118.0} & \demph{15.4} & \demph{116.3} & \demph{15.1} & \demph{120.2} & \demph{14.5} & \demph{117.3} & \demph{15.0} & \demph{42.6}  & $-$ & $-$& \demph{145.5} \\
           \rowcolor{LightCyan}
       \demph{Beit-3~\cite{beit3}} & \demph{35M+} & \demph{1.9B}& \demph{$-$}&\demph{$-$}&\demph{$-$}&\demph{$-$}&\demph{$-$}&\demph{$-$}&\demph{$-$}&\demph{$-$}&\demph{44.1}&\demph{32.4}&\demph{25.4} & \demph{147.6} \\
	\bottomrule
	\end{tabular}
 	}
% \vspace{-2ex}
	\vspace{-1ex}
	\label{tab:caption}
\end{table*}		

%% file: Tables/tab4_vqa_nlvr2.tex
\begin{table}[h]
	\centering	
    	\caption
	{
		\textbf{Comparison with state-of-the-art methods on VQA and NLVR$^2$.} 
		Para. is short for parameters.
		Notice that VinVL~\cite{vinvl} uses larger vision backbone and object feature from faster-rcnn.
		ALBEF~\cite{albef} performs an extra pre-training step for NLVR$^2$.
	}
    	\resizebox{\columnwidth}{!}{%
	\begin{tabular}	{lll|llll}
		\toprule	 	
	 \multirow[t]{2}{*}{\bf Method}& \multirow[t]{2}{*}{\bf \#Images}&
	 \bf Para. & \multicolumn{2}{c}{\bf VQA} & \multicolumn{2}{c}{\bf NLVR$^2$} \\
	  &&& test-dev & test-std & dev & test-P \\
	  \midrule
	  UNITER \cite{uniter}& 4M &155M& 72.70 & 72.91 & 77.18 & 77.85 \\
	   OSCAR \cite{oscar}& 4M &155M& 73.16 & 73.44 & 78.07 & 78.36 \\
	   %VILLA& 4M & 73.59 & 73.67 & 78.39 & 79.30\\
	   UNIMO~\cite{unimo} & 5.6M &307M& 75.06 & 75.27 & - & - \\
	   VinVL$_{L}$ \cite{vinvl}&5.6M&347M&\bf 76.52&\bf 76.60&\bf 82.67&\bf 83.98\\
	   ViLT \cite{vilt} & 4M &87M& 70.33 & -& 74.41 & 74.57 \\
  	   \bf \ModelName-ViLT   &4M &87M& 72.13$_{+1.8}$& 74.36 & 76.52$_{+2.1}$ & 77.83$_{+3.3}$\\
   	   BLIP $\dag$ \cite{blip} & 4M &220M& 73.92&74.13 & 77.52  & 77.63 \\
	   \bf \ModelName-BLIP  & 4M &220M&  76.02$_{+2.1}$ & 76.18$_{+2.0}$ &  80.73$_{+3.2}$ & 81.24$_{+3.8}$  \\
	   \midrule
   	   ALBEF \cite{albef} & 14M &210M& 75.84 & 76.04 & 82.55 & 83.14\\ 
	   BLIP \cite{blip} & 14M &220M& 77.54 & 77.62 & 82.67 & 82.30 \\
   	   \bf \ModelName-BLIP   & 14M & 220M& \bf 78.44$_{+2.9}$ & \bf 78.33$_{+1.7}$ & \bf 84.55$_{+1.9}$ & 83.17$_{+0.9}$  \\
	   \midrule
	   	   \rowcolor{LightCyan}
	   \demph{SimVLM} \cite{simvlm} & \demph{1.8B} &\demph{1.2B}& \demph{77.87} & \demph{78.14} & \demph{81.72} & \demph{81.77} \\
	     \rowcolor{LightCyan}
  	   \demph{GIT} \cite{git} &\demph{0.8B} & \demph{0.7B}& - & \demph{78.81}&-&-\\
      \rowcolor{LightCyan} 
\demph{CoCa} \cite{coco} & \demph{3B}&\demph{2.1B}&\demph{84.2 }&\demph{84.0}&\demph{86.1}&\demph{87.0}\\
  	   % 	     \rowcolor{LightCyan}
  	   % \demph{Beit-3} \cite{beit3} &\demph{35M+} &\demph{1.9B}& \demph{84.19} & \demph{84.03} & \demph{91.51}&\demph{92.58}\\
	\bottomrule
	\end{tabular}
	}
	\label{tab:vqa_nlvr}
\end{table}

%% file: Tables/tab5_msrvtt_retrieval.tex
\begin{table}[t]
\centering
    \caption
	{
	\textbf{Comparisons with state-of-the-art methods for text-to-\textbf{video} retrieval on the 1k test split of the MSRVTT dataset.
	}}
\scalebox{.8\GlobalTableRescale}
{
		\begin{tabular}	{l|cccc}
		\toprule
		\bf Method & \bf R@1$\uparrow$ & \bf R@5$\uparrow$ & \bf R@10$\uparrow$ & \bf MdR$\downarrow$\\
		\midrule
		ActBERT \cite{actbert} & 8.6 & 23.4 & 33.1 & 36.0\\
        MIL-NCE \cite{milnce} & 9.9 & 24.0 & 32.4 & 29.5\\ 
        %VideoCLIP& 10.4 & 22.2 & 30.0 & -\\
        Frozen-in-time \cite{frozen}& 18.7 & 39.5 &51.6& 10.0 \\ 
        OA-Trans \cite{oatrans} & 23.4 & 47.5 & 55.6 & 8.0 \\
        %FiT& 18.7 & 39.5 & 51.6 & 10 \\ 
        \midrule
        \ModelName-ViLT & \bf 27.9 &\bf 52.5 & \bf 56.3 & \bf 7.0 \\
        %\ModelName-BLIP & \\
        \bottomrule
		\end{tabular}
 		}
 	\vspace{-2ex}
	%\vspace{-1ex}
	\label{tab:zsl_video_retrieval}
\end{table}		

%% file: Tables/tab10_abl_architevtures.tex
\begin{table}[t]
\centering
\small
\caption{
\textbf{The ablation on different architectures under 4M setting.
}
We report the i2t and t2i results on MSCOCO (5K test set).
As we do not used object detector in downstream tasks, \ModelName is 20 times faster than object-feature based model.
}
\resizebox{.9\linewidth}{!}{%
\begin{tabular}{ll|ccccccl}
\toprule
\bf Method & \bf Time &\multicolumn{7}{c}{\bf MSCOCO (5K test set)} \\
 && \multicolumn{3}{c}{Image $\rightarrow$ Text} & \multicolumn{3}{c}{Text $\rightarrow$ Image}\\
 \cmidrule(lr){3-5} \cmidrule(lr){6-8}
 && R@1 & R@5 & R@10 & R@1 & R@5 & R@10 & Avg \\
\midrule
&&\multicolumn{7}{c}{\textit{One-stream Models}} \\
ViLT \cite{vilt}&$\sim$15& 61.8 & 86.2 & 92.6 & 41.3 & 72.0 & 82.5 & 72.7 \\
\bf  \ModelName-ViLT &$\sim$15& \bf 67.1 &\bf  90.5 &\bf  94.3 & \bf 45.3 &\bf  79.1 &\bf  88.4 & \bf 77.5$_{+4.8}$ \\
\cdashline{1-9}[0.8pt/2pt]
&&\multicolumn{7}{c}{\textit{Dual-stream Models}} \\
CLIP$\dag$ \cite{CLIP} &$\sim$27&64.9 & 83.2 & 90.1 & 50.4 & 76.3& 84.7 & 74.9 \\
\bf \ModelName-CLIP &$\sim$27&\bf 68.3 &\bf  86.4 & \bf 92.7 &\bf 54.1 &\bf 80.1 &\bf 86.8 & \bf 78.1$_{+3.2}$  \\
\cdashline{1-9}[0.8pt/2pt]
&&\multicolumn{7}{c}{\textit{Dual-stream + Fusion encoder Models}} \\
BLIP $\dag$ \cite{blip} &$\sim$33& 75.2 & 93.3 &  96.3  &  57.4 &  82.1 & 89.5 & 82.3 \\
 \bf \ModelName-BLIP &$\sim$33& \bf 77.6 &\bf 94.2 &\bf  97.0  &\bf 59.4 &\bf 83.4 &\bf 90.4 &\bf 83.7$_{+1.5}$  \\
 \midrule
 &&\multicolumn{7}{c}{\textit{Object-feature Based Models}} \\
 \demph{VinVL~\cite{vinvl}}&\demph{$\sim$650}& \demph{74.9} & \demph{92.6} & \demph{96.3}  & \demph{58.1} & \demph{83.2} & \demph{90.1} & \demph{82.5} \\
\bottomrule
\end{tabular}
}
\label{tbl:abl_architectures}
\end{table}

%% file: Tables/tab9_pretext_vs_prompt.tex
\begin{table}
\footnotesize
\centering
\caption{
\textbf{Text prompt vs. additional pretext head.
}
% We observe text prompt leads to better result for all tasks.
The last column is COCO captioning task.
}
\scalebox{.8\GlobalTableRescale}
{
\begin{tabular}{l|llll}
\toprule
\bf Method & \bf COCO &\bf F30K &\bf NLVR&\bf Captioning\\
& TR@1  & TR@1& Acc(\%)& CIDER \\
%\shline
\midrule
\demph{Baseline}&\demph{70.6} & \demph{53.4} & \demph{76.1} & \demph{121.2} \\
\midrule
Pretext & 72.3 \greenp{1.7$\uparrow$} & 54.7 \greenp{2.3$\uparrow$}  & 76.9 \greenp{0.8$\uparrow$}  & 123.5 \greenp{2.3$\uparrow$}  \\
Prompt & \bf 73.2 \greenp{2.6$\uparrow$} & \bf 55.4 \greenp{2.0$\uparrow$}  & \bf 77.9 \greenp{1.8$\uparrow$}  & \bf 127.2 \greenp{6.0$\uparrow$} \\
\bottomrule
\end{tabular}
}
\label{tab:pretext_vs_prompt}
%\vspace{-2em}
\end{table}

%% file: Tables/tab6_prompt_abl.tex
\begin{table}
\footnotesize
\centering
\caption{
\textbf{Case study of text prompt on image-text retrieval.
}
A single-word change in prompt could yield a drastic difference.
O is short for object and P is short for position.
}
\scalebox{.9\GlobalTableRescale}
{
\begin{tabular}{l|ll}
\toprule
\bf Prompt & \bf TR@1 &\bf IR@1 \\
%\hline
\midrule
\demph{Baseline}&\demph{70.6} & \demph{53.4} \\
\hline
The [\textcolor{ocolor}{O}] is in the block [\textcolor{pcolor}{P}]. & 72.7 \greenp{2.1$\uparrow$} & 54.1 \greenp{0.7$\uparrow$}   \\
The block [\textcolor{pcolor}{P}] looks like [\textcolor{ocolor}{O}]. & 73.3 \greenp{2.7$\uparrow$} & 53.9 \greenp{0.5$\uparrow$} \\
The [\textcolor{ocolor}{O}] is in which block? In [\textcolor{pcolor}{P}].& 72.3 \greenp{1.7$\uparrow$} & 54.9 \greenp{1.5$\uparrow$}  \\
The [\textcolor{ocolor}{O}] is located in block [\textcolor{pcolor}{P}]. & 72.3 \greenp{1.7$\uparrow$} & 54.2 \greenp{0.8$\uparrow$} \\
(X1, Y1, W, H) has a [\textcolor{ocolor}{O}]. & 72.5 \greenp{1.9$\uparrow$} & 54.3 \greenp{0.9$\uparrow$} \\
The block in [NP] has a [\textcolor{ocolor}{O}]. & 73.0 \greenp{2.4$\uparrow$} &  55.1 \greenp{1.7$\uparrow$} \\
The block [\textcolor{pcolor}{P}] has a [\textcolor{ocolor}{O}]. & \bf 73.2 \greenp{2.6$\uparrow$} & \bf 55.4 \greenp{2.0$\uparrow$} \\
\midrule
Mixed & 72.3 \greenp{1.7$\uparrow$} & 54.7 \greenp{1.2$\uparrow$} \\
\bottomrule
\end{tabular}
}
\label{tab:abl_prompt_design}
%\vspace{-2em}
\end{table}

%% file: Tables/tab7_object_pred_var.tex
\begin{table}
\footnotesize
\centering
\caption{
\textbf{The position information is essential for prompt design.}
Different variations of object prediction prompt design and evaluate on coco retrieval.  
}
\scalebox{.8\GlobalTableRescale}
{
\begin{tabular}{ccc|ll}
\toprule
\bf Object Tags & \bf Prompt &\bf Position &\bf TR@1& \bf IR@1 \\
%\shline
\midrule
-&-&-&\demph{70.6}& \demph{53.4} \\%
\midrule
\textcolor{green}{\checkmark}& & &70.2 \redp{0.4$\downarrow$} & 52.7 \redp{0.7$\downarrow$} \\
\textcolor{green}{\checkmark}&\textcolor{green}{\checkmark} && 70.3 \redp{0.3$\downarrow$} &52.9 \redp{0.5$\downarrow$}\\
\textcolor{green}{\checkmark}& &\textcolor{green}{\checkmark}& 70.8 \greenp{0.3$\downarrow$} &52.4 \redp{1.0$\downarrow$}\\
\textcolor{green}{\checkmark}&\textcolor{green}{\checkmark}&\textcolor{green}{\checkmark}& \bf 73.3 \greenp{2.7$\uparrow$} & \bf 55.4 \greenp{2.0$\uparrow$} \\
\bottomrule
\end{tabular}
}
\label{tab:abl_object_pred}
\vspace{-2em}
\end{table}

%% file: Tables/tab8_remove_detector.tex
\begin{table}[!t]
\footnotesize
\centering
\caption{
\textbf{The different ways to get grid pseudo label and its corresponding running time.} 
We report the image-to-text retrieval result on the COCO dataset for reference.
}
\scalebox{.8\GlobalTableRescale}
{
\begin{tabular}{ll|ccc}
\toprule
\bf Method & \bf Time &\bf R@1& \bf R@5& \bf R@10 \\
%\shline
\midrule
\demph{baseline} & - & \demph{70.6} & \demph{91.3} & \demph{95.4}  \\
\midrule
Faster-RCNN (ResNet101) & 10d & 72.7 & 91.8 & 95.7  \\
Faster-RCNN (ResNeXt152) & 14d & \bf 73.3 & \bf 92.0 & 96.1  \\
CLIP Similarity & 8h & 72.9&\bf 92.0 & \bf 96.6 \\
\bottomrule
\end{tabular}
}
\vspace{-1.5em}
\label{tab:remove_detector}
\end{table}

%% file: Figures/4_Abl_Patch_Num.tex
\begin{figure}[h]
        \centering
        \includegraphics[width=.8\columnwidth]{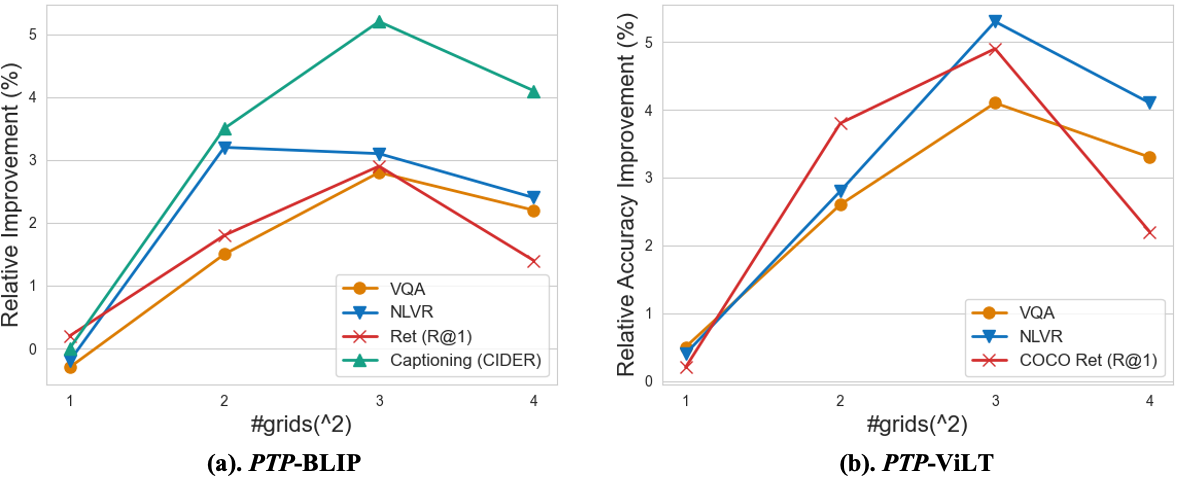}
        \vspace{-1em}
        \caption{%
        \textbf{
        The relation between the number of blocks and the relative accuracy improvement.
        }
        We explore two baselines and show the improvements over four different tasks.
        }
        \vspace{-1em}
        \label{fig:ablation_patch_num}
\end{figure}

%% file: Figures/4_Block_Pred_Visualize.tex
\begin{figure}[!t]
        \centering
        \includegraphics[width=.88\linewidth]{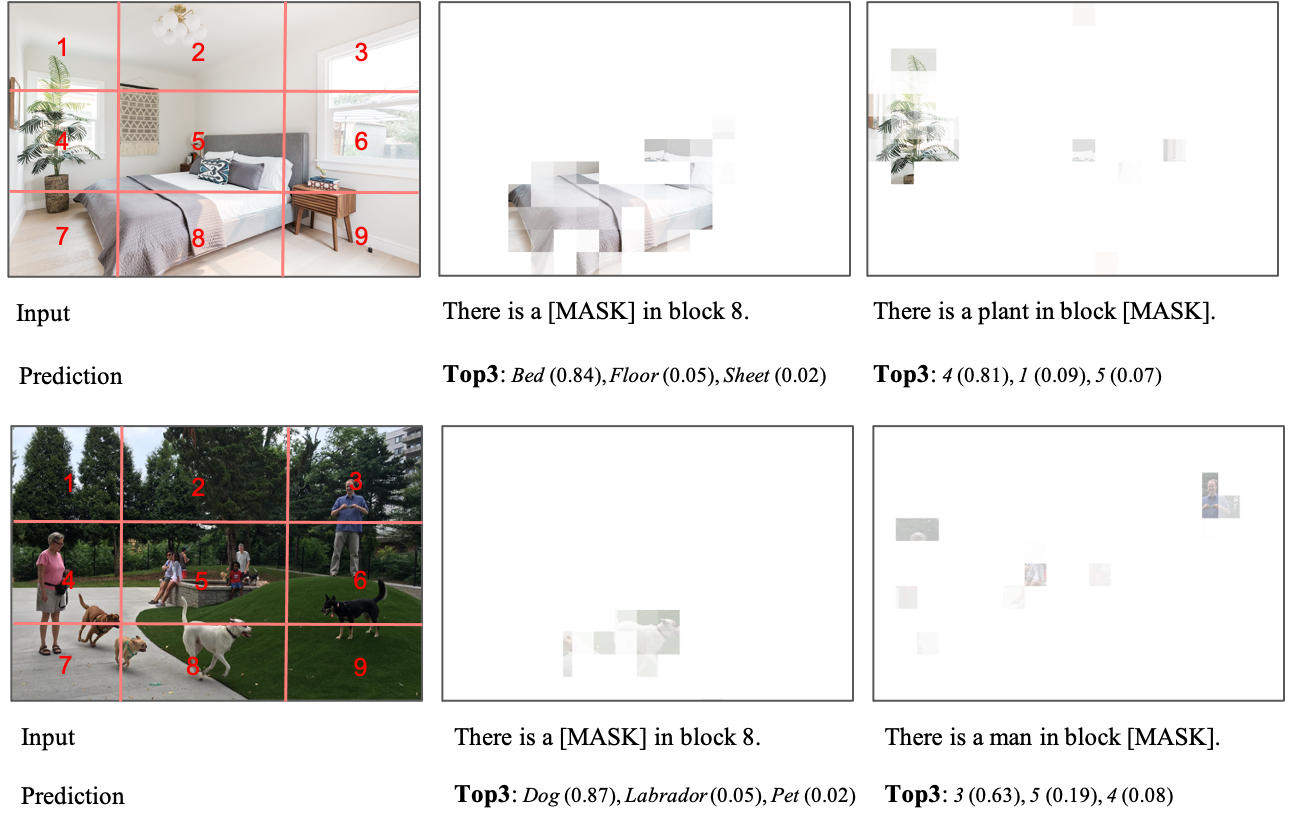}
        \vspace{-1em}
        \caption{%
        \textbf{The full-in-the-blank task evaluation.}
        We ask the model to predict \textit{what objects are contained in given block} and \textit{predict which blocks contain specific object}.
        %and its corresponding attention map
        }
        \vspace{-1.3em}
        \label{fig:block_pred}
\end{figure}

%% file: Figures/4-Token_Visualize.tex
\begin{figure}[h]
        \centering
        \includegraphics[width=\linewidth]{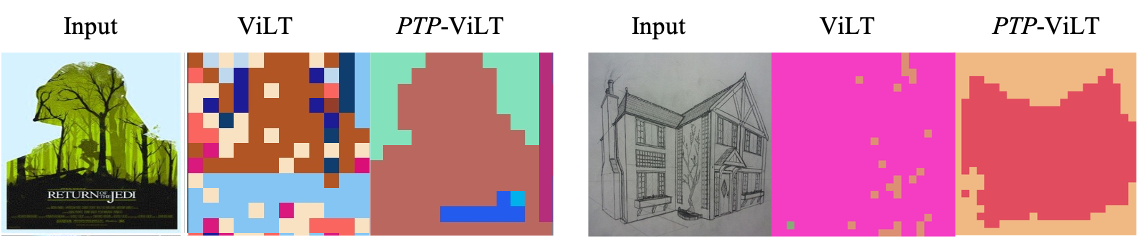}
        \caption{%
        \textbf{Token cluster visualization.}
        We train ViLT and \ModelName-ViLT with ViT-B/32 model on CC3M train set.
        We show the token cluster result with KMeans algorithm from CC3M test set \cite{cc3m}.
        \ModelName-ViLT shows preferable clusters. 
        %See Section \ref{sec:4_visulaize} for more details.
        }
        \vspace{-1em}
        \label{fig:token_visualize}
\end{figure}

%% file: Supplementary/tab1_dataset_statistic.tex
\begin{table}[ht]
\small
\centering	
\scalebox{.9\GlobalTableRescale}
{
\begin{tabular}	{l | l | l |  l | l }
\toprule
 & \bf Dataset & \# \bf Images  & \# \bf Captions & \# \bf BBox \\
\midrule
 \multirow{4}{*}{4M} & COCO & 0.11M & 0.55M & 0.11M \\
 	 & Visual Genome & 0.10M & - & 0.10M  \\
 & SBU & 0.86M & 0.86M & - \\
 & CC-3M & 2.8M & 2.8M & 2.69M \\
 \midrule
 \multirow{2}{*}{14M} & 4M & 4.0M & 5.1M & 2.9M \\
 & CC-12M & 10.2M & 10.2M & 7M \\
 \bottomrule
\end{tabular}
}
\caption{Statistics of the pre-training datasets. 
}
\label{tab:data}
\end{table}

%% file: Supplementary/tab3_ft_details.tex
\begin{table}[!t]
\small
\centering
\resizebox{.95\linewidth}{!}{
\begin{tabular}{@{}l|cccc}
    \toprule
    \bf Task & \bf VQA & \multicolumn{1}{c}{\bf Retrieval}& \bf NLVR2 & \multicolumn{1}{c}{ \bf Captioning} \\
    \cmidrule(r){1-1} \cmidrule(l){2-5}
    Optimizer        & \multicolumn{4}{c}{AdamW with Weight Decay} \\
    Gradient clip         & \multicolumn{4}{c}{1.0} \\
    LR decay schedule     & \multicolumn{4}{c}{Cosine Schedule Decaying to Zero} \\
    Weight decay rate     & \multicolumn{4}{c}{0.05}   \\
    RandAugment           & 2,5       &  2,5   & 2,5   & 2,5   \\
    Train epochs         &  10          &  6     &    5  & 5  \\
    Train batch size      & 64         & 24     & 128    & 16 \\
    LR    & 2e-5  & 1e-5   & 3e-5    & 1e-5 \\
    %Warm-up steps              & 1000 & 1000    & 1000 & 1000  & 1000    \\
    \bottomrule
\end{tabular}
}
\caption{Hyper-parameters used in the multimodal experiments for BLIP baseline.}
\vspace{-1em}
\label{tabs:detailed_vl}
\end{table}

\begin{table}[!h]
\small
\centering
\resizebox{.8\linewidth}{!}{
\begin{tabular}{@{}l|cccc}
    \toprule
    \bf Task & \bf VQA & \multicolumn{2}{c}{\bf Retrieval}& \bf NLVR2\\
     Dataset & VQAV2 & COCO  & F30K & NLVR2 \\
    \cmidrule(r){1-1} \cmidrule(l){2-5}
    Optimizer        & \multicolumn{4}{c}{AdamW with Weight Decay} \\
    Gradient clip         & \multicolumn{4}{c}{1.0} \\
    LR decay schedule     & \multicolumn{4}{c}{Cosine Schedule Decaying to Zero} \\
    RandAugment           & \multicolumn{4}{c}{2,9} \\
    Weight decay rate     & \multicolumn{4}{c}{0.05} \\
    Train epochs         &  10     &  10    &  5     &    10 \\
    Train batch size      & 256      & 256   & 256    & 128   \\
    LR    & 1e-4 & 3e-4 & 1e-4 & 1e-4   \\
    Warm-up steps   & 1500 & 2500 & 1000 & 500  \\
    \bottomrule
\end{tabular}
}
\caption{Hyper-parameters used in the multimodal experiments for ViLT baseline.}
\label{tabs:detailed_vl_vilt}
\end{table}

%% file: Supplementary/tab2_more_prompt.tex
\begin{table*}[ht]
\footnotesize
\centering
\scalebox{.9\GlobalTableRescale}
{
\begin{tabular}{l|ccc|ccccc}
\toprule
\bf Prompt & \bf Multipy& \bf Multipy& \bf Prompt & \multicolumn{2}{c}{\bf COCO Retrieval} &\bf NLVR & \bf COCO Captioning  \\
& \bf Position  & \bf Tags   &  & TR@1 & IR@1 & Acc & CiDER \\
%\hline
\midrule
\demph{Baseline}&-&-&-&\demph{70.6} & \demph{53.4} & \demph{76.0} & \demph{122.6} \\
\hline
The object in region [\textcolor{pcolor}{P}] looks like [\textcolor{ocolor}{O}]. &&&\checkmark& 72.5 \greenp{1.9$\uparrow$} & 54.3 \greenp{0.9$\uparrow$} & 77.8 \greenp{1.8$\uparrow$} & 127.4 \greenp{4.8$\uparrow$}\\
The block [\textcolor{pcolor}{P}] has objects [\textcolor{ocolor}{O$_1$}], [\textcolor{ocolor}{O$_2$}], [\textcolor{ocolor}{O$_3$}]. & & \checkmark &  \checkmark &71.9 \greenp{0.9$\uparrow$} &54.7 \greenp{0.9$\uparrow$} &76.8 \greenp{0.9$\uparrow$} & 124.5 \greenp{1.9$\uparrow$}  \\
The [\textcolor{ocolor}{O}] is located in which region? In [\textcolor{pcolor}{P$_1$}], [\textcolor{pcolor}{P$_2$}] and [\textcolor{pcolor}{P$_3$}]. &\checkmark&&\checkmark&70.7 \greenp{0.1$\uparrow$} &53.6 \greenp{0.2$\uparrow$} &77.1 \greenp{1.1$\uparrow$} & 125.2 \greenp{2.6$\uparrow$}  \\
\bottomrule
\end{tabular}
}
\caption{
\textbf{
Other variations of text prompt.
}
[\textcolor{ocolor}{O}] is short for object and [\textcolor{pcolor}{P}] is short for position.
}
\label{tab:more_prompt}
%\vspace{-2em}
\end{table*}

%% file: Supplementary/figure1_block_mask.tex
\begin{figure*}[ht]
        \centering
        \includegraphics[width=\linewidth]{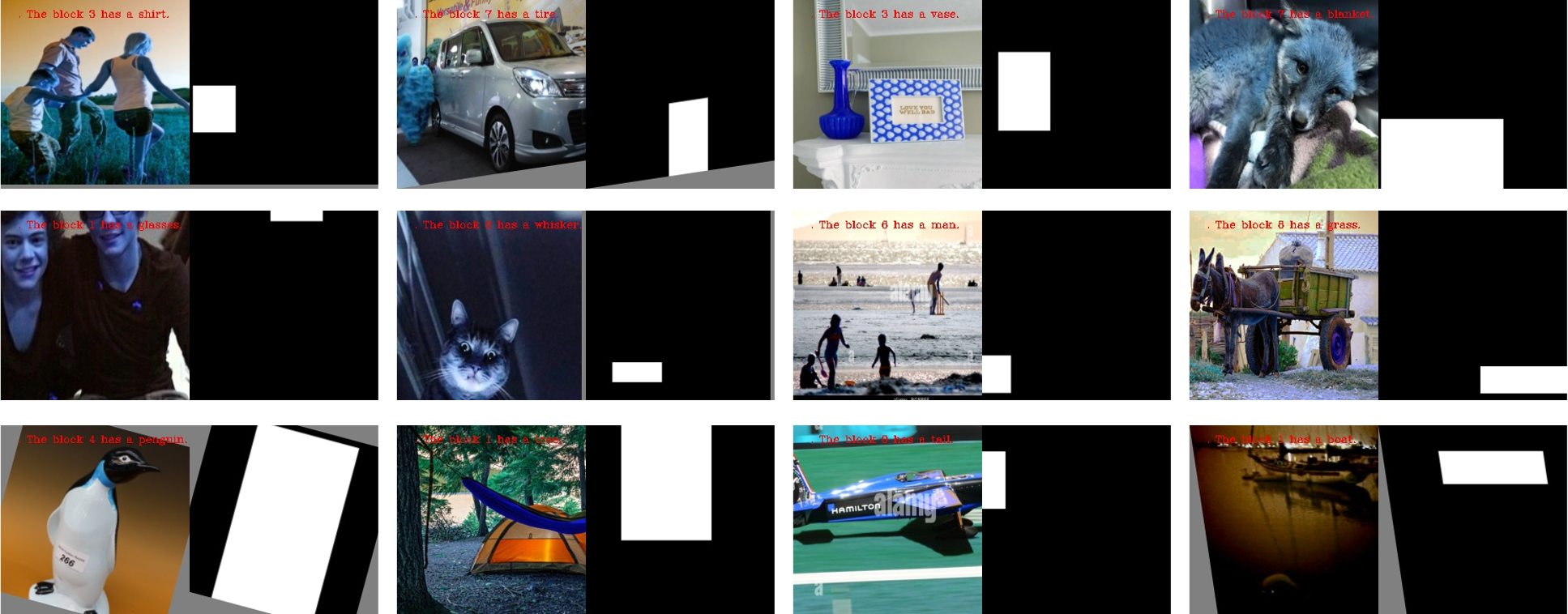}
        \caption{%
        \textbf{Our text prompt (in red color) and its corresponding bounding box's mask.}
        The block index is from 0 to 8.
        }
        \label{fig:block_mask}
\end{figure*}

%% file: Supplementary/figure2_vary_objects.tex
\begin{figure}[h]
        \centering
        \includegraphics[width=\linewidth]{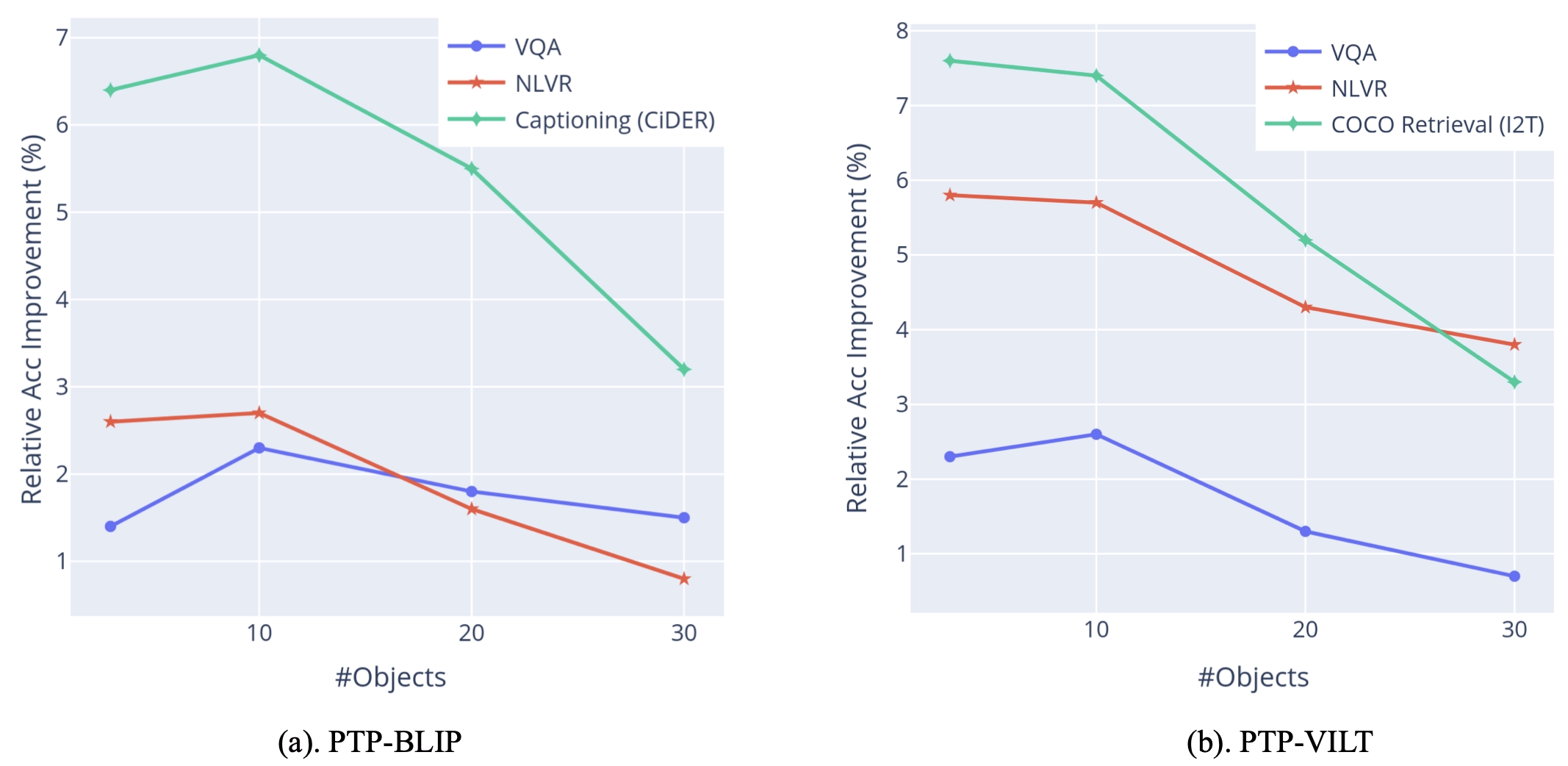}
        \caption{%
        \textbf{We varying the number of selecting objects from $3$ to $30$.}
        We report the result on downstream tasks over BLIP and ViLT baselines.
        }
        \label{fig:vary_object}
\end{figure}

%% file: Supplementary/tab4_part_object.tex
\begin{table}
\footnotesize
\centering
\scalebox{.9\GlobalTableRescale}
{
\begin{tabular}{l|ccccc}
\toprule
\bf Method & \multicolumn{2}{c}{\bf COCO Retrieval} &\bf NLVR & \bf COCO Captioning  \\
&  TR@1 & IR@1 & Acc & CiDER \\
%\hline
\midrule
\demph{0\%}&\demph{79.5} & \demph{62.4} & \demph{80.5} & \demph{129.5} \\
\hline
19.3\%&82.2  & 65.1  & 81.4  & 140.1 \\
68.6\%&83.7 &68.4  &82.9  & 143.6  \\
\bottomrule
\end{tabular}
}
\caption{
\textbf{
Part samples with position information.
}
Under 14M setting, we test the result with different amount of pre-training samples with objects.
}
\label{tab:part_object}
%\vspace{-2em}
\end{table}

%% file: Supplementary/figure3_downstream_visualization.tex
\begin{figure}[ht]
        \centering
        \includegraphics[width=\linewidth]{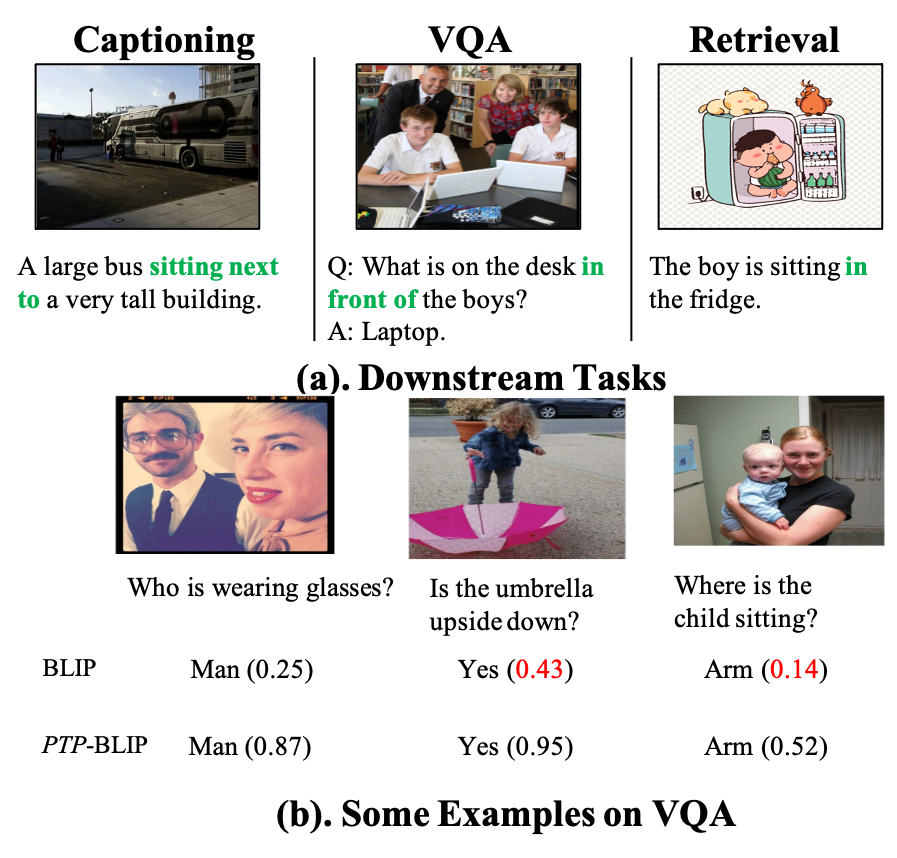}
        % \vspace{0.8em}
        \caption{
        Mainstream downstream tasks all requires position information.
        }
        \label{fig:samples_visualization}
\end{figure}